\documentclass[final]{clv2025}

\jvol{vv}
\jnum{nn}
\jyear{2025}

\dochead{Short paper} 

\pageonefooter{Action editor: \{action editor name\}. Submission received: DD Month YYYY; revised version received: DD Month YYYY; accepted for publication: DD Month YYYY.}

\usepackage{graphicx} 
\usepackage{xcolor}
\usepackage{amsmath} 
\usepackage{amssymb} 
\usepackage{amsthm}
\DeclareMathOperator{\E}{\mathbb{E}}
\usepackage{bbm} 

\newcommand{\commentinequation}[1]{\mbox{~~~(#1)}} 

\usepackage{tabularx}
\usepackage{lscape}
\usepackage{booktabs}

\usepackage{etoolbox}

\newtoggle{publish}
\togglefalse{publish}

\newtoggle{PSUD}
\togglefalse{PSUD}

\input{table_of_contents} 

\newtheorem{property}[theorem]{Property}

\runningtitle{Parsing evaluation without a gold standard}
\runningauthor{Ferrer-i-Cancho, Hobaiter, Gustison \& Bergman}

\begin{document}

\allowdisplaybreaks

\title{On the feasibility of dependency parsing of non-human sequences without a gold standard. Is evaluation possible in other species? }


\author{Ramon Ferrer-i-Cancho\thanks{Corresponding author}$^{,1}$, Catherine Hobaiter$^{2}$, Thore Bergman$^{3}$ Morgan Gustison$^{4}$}

\affilblock{
    \affil{Universitat Politècnica de Catalunya, Department of Computer Science \\\quad \email{rferrericancho@cs.upc.edu}}
    \affil{University of St Andrews, School of Psychology and Neuroscience \\\quad \email{clh42@st-andrews.ac.uk}}
    \affil{University of Michigan, Departments of Psychology and Ecology and Evolutionary Biology \\\quad \email{thore@umich.edu}}    
    \affil{Western University, Department of Psychology \\\quad \email{mgustiso@uwo.ca}}    
}

\maketitle

\begin{abstract}
Dependency parsing consists of finding a tree representation for a sequence. Unsupervised dependency parsing aims to develop parsing methods without a gold standard during model training. 
In human languages, an unsupervised parser can be evaluated because some gold standard is usually available or can be created. For other species, a gold standard is unknown. Thus one may conclude that it is impossible to determine the accuracy of an unsupervised parser and, consequently, dependency parsing is unfeasible in other species. However, here we apply recent advances in network science to demonstrate that the proportion of correct edges retrieved by a parser must be high for the sequences of vocalizations or gestures that non-human primates produce due to the fast decay of the sequence length distribution. In contrast, human language sequences lack that property. Therefore, evaluation without a gold standard is feasible in non-human primates but a hard problem in humans. 
\end{abstract}

\iftoggle{publish}{}
{
}

\section{Introduction}

\begin{figure}
    \centering
    \includegraphics[width=0.95\linewidth]{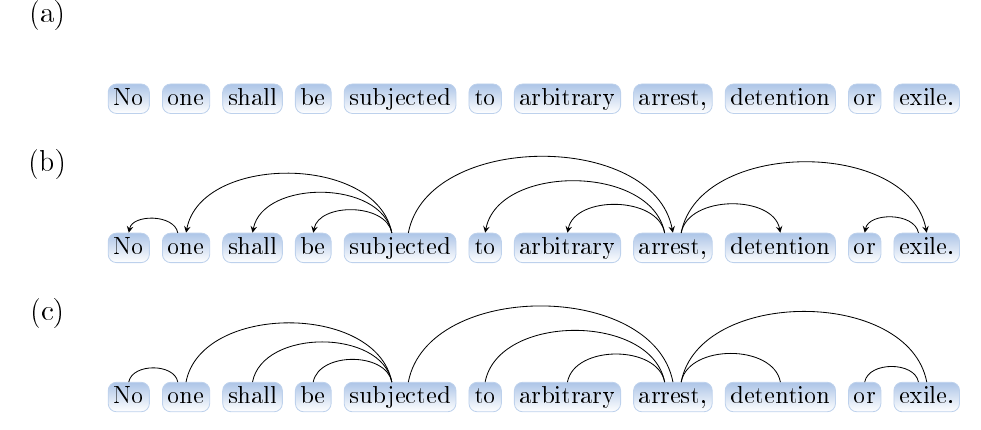}
    \caption{\label{fig:Article9_UDHR} 
    (a) The raw sequence of words making the sentence.
    (b) The syntactic dependency parse of Article 9 of the Universal Declaration of Human Rights by Stanza 1.10.1 \url{http://stanza.run/}.
    (c) The same parse as in (b) after removing dependency direction. }
\end{figure}

The syntactic structure of a sentence (Figure \ref{fig:Article9_UDHR} (a)) can be represented by a {\em syntactic dependency structure,} namely a rooted tree where vertices are words and arcs indicate syntactic dependencies between words as in Figure \ref{fig:Article9_UDHR} (b) \cite{Melcuk1988}. A syntactic dependency treebank, hereafter a {\em treebank}, is a collection of sentences and their dependency structure. Dependency parsing, hereafter {\em parsing}, is finding the tree representation of a sentence automatically, and a {\em parse} is the tree representation retrieved by a {\em parser} for a sentence \cite{Marecek2012a,Carroll2014a}.
Supervised parsers learn to parse by means of a {\em gold standard,} usually a treebank with the correct or desired dependency structure for each sentence (Figure \ref{fig:Article9_UDHR} (b)). 
Supervised learning consists of fitting a model by means of a treebank as the gold standard. {\em Unsupervised parsers} learn to parse without a treebank, namely they learn to parse just from a collection of raw sentences, that is just the word sequence (as in Figure \ref{fig:Article9_UDHR} (a)) or the word sequence plus part-of-speech information  \cite{Han2020a}. {\em Unsupervised learning} consists of fitting a model without a gold standard.
{\em Evaluation} consists of comparing the parses against a gold standard treebank. Typical evaluation scores are the proportion of trees correctly retrieved or the proportion of dependencies correctly retrieved \cite{Carroll2014a,Han2020a}. 

In many languages, the evaluation of a parser is straightforward because treebanks are already available. In other languages, evaluation is not possible because treebanks are not forthcoming (the so-called ``low-resource'' languages), their development is technically costly \cite{Scheinerman2012a} or the language itself poses theoretical challenges in how to define the structure of sentences (see for instance \citet{Vasquez2018a} or \citet{Martin-Rodriguez2022a}). For other species, a gold standard is simply not forthcoming, representing a serious challenge for mainstream linguistics, which is human-language based. Although an unsupervised parser can be used to retrieve dependency structures from the sequences that other species produce, the hard problem is the evaluation: the parser will produce dependency structures but how can one know if they are correct? Thus, one may conclude that one cannot evaluate the performance of a parser in other species, which, in turn, may lead one to  conclude that accurately parsing the communication of other species is not viable. Unsupervised parsing could be applied to other species but would only serve to speculate on the range of possible structures that other species may have. 

In spite of the considerations above, other primates produce sequences that are rather short \cite{Gustison2016a,Girard-Buttoz2022a,Mielke2024a} and the distribution of sequence lengths decays exponentially across species (e.g., geladas: \citet{Gustison2016a}; chimpanzees: \citet{Girard-Buttoz2022a}). Here we apply recent advances in network science \cite{London2023a} to demonstrate that the proportion of correct edges retrieved by a parser must be high for the sequences of vocalizations or gestures that non-human primates produce due to the fast decay of the sequence length distribution. In contrast, human language sentences lack that property. Our core argument is as follows. First, we neglect dependency direction leaving just edges without direction (Figure \ref{fig:Article9_UDHR} (c)), a common practice in the evaluation of unsupervised parsers \cite{Marecek2012a,Han2020a}. Second, we consider a  random parser for a sequence of length $n$ such that $n \geq 2$. 
That parser selects the correct tree at random over all possible trees of $n$ vertices. We will show that the expected proportion of edges that are correctly identified by the random parser is $2/n$. If $n = 2$, the parser has $100\%$ maximum accuracy. If $n = 3$, its expected accuracy is $66\%$, if $n=4$ then $25\%$, ... Therefore, if sequence lengths are biased towards low values, the random parser will have high accuracy. Therefore, any good-enough unsupervised parser, namely any parser theoretically able to improve over a random parser by exploiting the statistical structure of the sequences \cite{Marecek2016a,Han2020a}, will achieve an accuracy that is superior to that of the random parser. Indeed, the sequences that non-human apes produce have some statistical structure \cite{Gustison2016a,Gustison2017a,Mielke2024a}. Therefore, even if we do not have a gold standard, we will always be able to provide a lower bound to the accuracy of any good-enough parser, and if sequence lengths are biased towards low values, we will be able to show that its performance must be high.

Here we aim to show that the performance of a good-enough parser on real-world vocal or gestural sequences produced by non-human primates is, indeed, expected to be high. The remainder of the article is organized as follows. Section \ref{sec:theory} investigates the theoretical ability of a random parser to find the right tree or to identify edges correctly in a collection of sequences. In particular, we 
present the expected value of the proportion of correct trees and the proportion of correct dependencies for three kinds of distribution of sequence lengths: the empirical distribution and two theoretical distributions, i.e. the uniform distribution and the geometric distribution. The latter approximates the distribution of vocal sequence lengths that geladas and chimpanzees produce \cite{Gustison2016a, Girard-Buttoz2022a}. 
Section \ref{sec:theory} investigates the impact of the parameters of the theoretical distributions above and shows that the performance of the random parser is high when sequences are short but decays as the average length of the sequences increases.
Section \ref{sec:methods} presents the datasets from humans and other primates that will be used to obtain the empirical distribution of sequence lengths or key parameters (e.g., the maximum sequence length) of that distribution. Such empirical distribution serves two purposes. First, an accurate estimation of the performance of the random parser. Second, providing support for the geometric distribution as a theoretical model for the sequences that non-human primates produce.  

Section \ref{sec:results} makes a series of contributions. 
First, Section \ref{sec:results} compares the shape of the distribution of sequence lengths in chimpanzee vocal and gestural sequences versus sentences in humans. It shows that only sequence lengths of chimpanzees can be approximated by a geometric distribution. Second, it estimates a lower bound of the performance of a random parser in 31 species of non-human primates. Third, it also provides an accurate estimate of the performance of the random parser in geladas, chimpanzees, and humans. The average proportion of correct edges retrieved per sequence by the random parser is $51\%$ for geladas and $>79\%$ for chimpanzees. If the analysis is restricted to sequences of more than two units, the performance drops to $41\%$ for geladas and to $>56\%$ for chimpanzees. In contrast, the accuracy of the random parser on human languages is about $27\%$ when restricted to sentences of length 10 and about $13\%$ for sentences of any length. Finally, Section \ref{sec:discussion} concludes that evaluation without a gold standard is feasible in geladas and chimpanzees but hard in humans. Section \ref{sec:discussion} also suggests that successful training of an unsupervised parser is likely for these species. 

\section{Theory}
\label{sec:theory}

The performance of a parser can be evaluated in different ways. First, by considering the direction of edges, as in Figure \ref{fig:Article9_UDHR} (b), or neglecting it as in Figure \ref{fig:Article9_UDHR} (c) \cite{Han2020a}. 
Here we disregard edge direction for simplicity and to avoid imposing {\em a priori} any notion of hierarchy or rootness in the dependency structures that other species produce \cite{Frank2018a}. 
Therefore, our target is the free tree of the sequence. Figure \ref{fig:Article9_UDHR} (c) shows the free tree of a sentence, that is obtained by removing link direction from the rooted tree (\ref{fig:Article9_UDHR} (b)).
Second, considering the number of correct trees retrieved (hard evaluation) or by considering the number of correct edges retrieved (soft evaluation) \cite{McDonald2005a}. Here we consider both aspects.

\subsection{Elementary definitions}

Here we wish to evaluate the performance of a parser on a collection of $S$ sequences. We assume that each sequence has a single correct dependency structure, the default assumption for the syntactic dependency structure of human language sentences. To simplify the analysis of the performance of the parser, we assume that the parser always retrieves a free tree when supplied a sequence. The parser may produce directly a free tree \cite{Yuret1998a} or it may produce a rooted tree that is then transformed to a free tree \cite{Han2020a}. Certain parsing methods may produce more than one free tree given a sequence. For instance, parsing methods that consist of extracting an optimal spanning tree from a graph $G$, may retrieve more than one free tree when there are tied weights in $G$ \cite{McDonald2005a}. If that happened, we assume that the parser eventually delivers just one of the tied trees chosen at random so as to simplify the theoretical analysis of the performance of that parser. That way, one does not need to distinguish the precision of the parser from its recall because they become the same.

We define $n_i$ as the length of the $i$-th sequence in a collection of $S$ sequences. In human languages, sequences can be made of words in spoken language or signs in sign languages. In other species, these sequences can be made of vocalizations \cite{McCowan1999, Girard-Buttoz2022a, Kershenbaum2016a}, gestures \cite{Liebal2004a,Hobaiter2011a,Mielke2024a} or other behavioral patterns \cite{Altmann1965a, Ferrer2005h}. Like humans, other species combine modalities when forming sequences \cite{Hobaiter2017a,Mine2024a}. 

The total number of tokens making the sequences is 
\begin{equation*}
T = \sum_{i=1}^S n_i,
\end{equation*}
where $n_i$ is the length (number of elements) of the $i$-th sequence. 
The sample of sequence lengths, i.e. 
\begin{equation}
\{n_1, n_2,..., n_i, ..., n_S\}, 
\label{eq:sample_of_sequence_lengths}
\end{equation} 
defines the empirical distribution of sequence lengths, which can be expressed compactly by means of $f(n)$, the number of sequences of length $n$. Then 
\begin{align*}
S & = \sum_{n=1}^{n_{max}} f(n) \\
T & = \sum_{n=1}^{n_{max}} n f(n),
\end{align*}
where $n_{max}$ is the maximum sequence length. 

We define ${\cal F}$ as the forest formed by the all the free trees that a parser has obtained for each of the sequences. Such a forest has $T$ vertices, $m = T - S$ edges and $S$ trees. 

\subsection{The evaluation scores}

As for hard evaluation, we calculate $P_c^t$, the fraction of correct trees in ${\cal F}$, that is   
\begin{equation*}
P_c^t = \frac{S_c}{S},    
\end{equation*}
where $S_c$ is the number of correct trees in ${\cal F}$. The subindex $c$ stands for correct and the subindex $t$ stands for trees.
$P_c^t$ is known as the complete metric \cite{McDonald2005a}.
As for soft evaluation, we calculate $P_c^e$, the fraction of correct edges in ${\cal F}$, that is 
\begin{equation*}
P_c^e = \frac{m_c}{m},
\end{equation*}
where $m_c$ is the number of edges in ${\cal F}$ that are correct. 
The subindex $c$ stands again for correct and the subindex $e$ stands for edges. $P_c^e$ corresponds to the so-called undirected dependency accuracy \cite{Klein2004a,Han2020a}. $P_c^e$ is equivalent to the precision and recall of a parser on retrieving correct dependencies. 

The ability of the parser to find correct edges can also be evaluated by means of the local performance of the parser on individual sequences. To that aim, we define, $P_{c,i}^e$ as the proportion of correct edges for the $i$-th tree of ${\cal F}$. 
We assume that $n_{min} > 1$ so that $P_c^e$ is defined on individual trees. 
$Q$, the average value of $P_{c,i}^e$ over all sequences is 
\begin{equation*}
Q = \frac{1}{S(n_{min})} \sum_{\substack{i=1 \\ n_i \geq n_{min}}}^{S} P_{c,i}^e. \\
\end{equation*}
We adopt the convention that $P_c^e = 1$ when $n = 1$ and $P_{c,i}^e = 1$ when $n_i = 1$.

We use $n$ to refer to the length of a sequence. 
Sequences can be of length one or greater but up to $n = 2$ if it is impossible for a parser to make a mistake. When $n = 1$ the structure is empty and when $n=2$ the structure must link the only two vertices. For that reason, the evaluation scores will have a parameter $n_{min}$ that defines the minimum sentence length taken into account. We define $S(n_{min})$ as the number of sequences of length $n_{min}$ or greater ($S = S(1)$).
Accordingly, $P_c^t$ is refined as 
\begin{equation}
P_c^t = \frac{S_c(n_{min})}{S(n_{min})},
\label{eq:proportion_of_correct_trees}
\end{equation}
where 
\begin{align}
S_c(n_{min}) & = \sum_{\substack{i=1 \\ n_i \geq n_{min}}}^S c_i \label{eq:number_of_correct_trees} 
\end{align}
and $c_i$ is a binary variable that indicates if the parse for the $i$-th sequence is correct ($c_i = 1$ if correct; $c_i = 0$ otherwise).

We define $m_{c,i}$ as the number of correct edges retrieved for the $i$-th sequence. Then $P_c^e$ is refined as
\begin{equation}
P_c^e = \frac{m_c(n_{min})}{m(n_{min})}, \label{eq:proportion_of_correct_edges}
\end{equation}
where
\begin{align}
m_c(n_{min}) & = \sum_{\substack{i=1 \\ n_i \geq n_{min}}}^S  m_{c,i} \label{eq:total_number_of_edges} \\
m(n_{min}) & = \sum_{\substack{i=1 \\ n_i \geq n_{min}}}^S (n_i-1) \\
  & = \sum_{n = n_{min}}^{n_{max}} f(n)(n - 1). \nonumber
\end{align}

\subsection{The expected value of the evaluation scores}

In order to be maximally agnostic about what we do not know, we assume all labeled trees of the same size have the same probability of being the correct tree.
We are interested in the expected value of the evaluation scores in two settings: (a) given only $S(n_{min})$ and the theoretical distribution of sequence lengths, namely their theoretical distribution is given and (b) given the empirical distribution of sequences lengths, namely the sample of sequence lengths (equation \ref{eq:sample_of_sequence_lengths}), or $f(n)$. 

As for (a), we assume that the $n_i$'s are identically distributed and define $p(n)$ as the probability that a sequence of the collection has length $n$.
We are interested in two single-parameter distributions. First, a uniform distribution with parameter $n_{max}$, i.e. 
\begin{equation}
p(n) = \left\{
          \begin{array}{ll}
          \frac{1}{n_{max}} & \mbox{if } 1 \leq n \leq n_{max}.\\
          0                 & \mbox{otherwise}.
          \end{array}
       \right.\label{eq:plain_uniform_distribution}   
\end{equation}
Second, a geometric distribution with parameter $q$, i.e. 
\begin{equation}
p(n) = \left\{
          \begin{array}{ll}
          q(1-q)^{n-1} & \mbox{if } n \geq 1 \\
          0            & \mbox{otherwise}.
          \end{array}
       \right.\label{eq:plain_geometric_distribution}   
\end{equation}
The uniform distribution serves as a baseline for any non-increasing distribution, e.g., the geometric distribution with $q>0$, which is specially useful for species where only the maximum sequence length is reported \cite{Girard-Buttoz2022a}. The choice of the geometric distribution is motivated by direct evidence in gelada vocalizations  \cite{Gustison2016a} and indirect evidence in chimpanzee vocalizations \cite{Girard-Buttoz2022a}. \footnote{\citet{Girard-Buttoz2022a} do not check explicitly if the distribution is geometric as \citet{Gustison2016a} but in Figure 1 of their article they show a linear decay in linear-log scale that can be approximated by a geometric distribution. }

$p(n | n \geq n_{min})$ is the probability that a sequence has length $n$ among sequences of length $n_{min}$ or greater. 
For the uniform distribution, one obtains a uniform distribution in the integer interval $[n_{min}, n_{max}]$, i.e. 
\begin{equation}
p(n | n \geq n_{min}) = \frac{1}{n_{max} - n_{min} + 1}.
\label{eq:uniform_distribution}
\end{equation}
For the geometric distribution, one obtains the well-known displaced geometric distribution \cite{Park2023a}
\begin{equation}
p(n | n \geq n_{min}) = 
        \left\{
          \begin{array}{ll}
          q (1-q)^{n - n_{min}} & \mbox{if } n \geq n_{min} \\
          0            & \mbox{otherwise}.
          \end{array}
        \right.
\label{eq:geometric_distribution}
\end{equation}
Finally, for the empirical distribution, 
\begin{equation}
p(n | n \geq n_{min}) = \frac{f(n)}{S(n_{min})}. \label{eq:empirical_sequence_length_probability}
\end{equation}

Suppose a function of $\phi(n)$. We define the expectation of $\phi$ as 
\begin{equation}
\E[\phi(n)] = \sum_{n = n_{min}}^{\nu_{max}} p(n | n \geq n_{min}), \phi(n). \label{eq:generic_expectation}
\end{equation}
where $\nu_{max} = \infty$ for the geometric distribution and $\nu_{max} = n_{max}$ for the uniform or the empirical distribution.

We analyze the ability of the random parser to find the correct tree by means of  $\E[P_c^t]$, the expected value of $P_c^t$.
The expected proportion of correct trees produced by the random parser is (Appendix \ref{app:expected_proportion_of_correct_trees})
\begin{equation}
\E[P_c^t] = \E[n^{2 - n}]. \label{eq:expected_proportion_of_correct_trees_main}
\end{equation}
We analyze the ability of the random parser to retrieve correct dependencies by means of $\E[Q]$ and $\E[P_c^e]$, the expected value of $P_c^e$ and $Q$, respectively. 
The expected average proportion of correct edges over sequences of same length produced by the random parser is (Appendix \ref{app:expected_proportion_of_correct_edges})
\begin{equation}
\E[Q]  = 2 \E\left[\frac{1}{n}\right] \label{eq:expected_proportion_of_intersecting_edges_over_trees_main}            
\end{equation}
while the expected overall proportion of correct edges produced by the random parser is (Appendix \ref{app:expected_proportion_of_correct_edges})
\begin{equation}
\E[P_c^e] = \frac{2 S(n_{min})}{m(n_{min})} \E\left[1 - \frac{1}{n}\right].
\label{eq:expected_proportion_of_intersecting_edges_over_edges_main}   
\end{equation}

Trivially, 
\begin{equation}
\E[Q] \geq \E[P_c^t]. \label{eq:guessing_edges_is_easier_than_guessing_trees}
\end{equation}
Notice $\E[Q] = \E[2/n] \geq \E[P_c^t] = \E[n^{2-n}]$ since $n^{3-n} \leq 2$ for $n \geq 1$. Besides, 
$\E[P_c^e]$ and $\E[Q]$ are directly related, since
\begin{align}
\E[P_c^e] & = \frac{S(n_{min})}{m(n_{min})}\left(2 -   2 \E\left[ \frac{1}{n} \right] 
\right) \commentinequation{expanding Eq. \ref{eq:expected_proportion_of_intersecting_edges_over_edges_main}} \nonumber \\
          & = \frac{S(n_{min})}{m(n_{min})}\left(2 -   \E[Q] \right) \label{eq:relationship_between_expected_performance} \\
\end{align}

The condition $\E[Q] \geq \E[P_c^e]$ becomes 
\begin{align}
\E[Q] & \geq \frac{2 S(n_{min})}{S(n_{min}) + m(n_{min})} \nonumber \\ 
      & = \frac{2 S(n_{min})}{S(n_{min}) + T(n_{min}) - S(n_{min})} \nonumber \\
      & = \frac{2 S(n_{min})}{T(n_{min})} = \E[Q]_*.   
\label{eq:expected_proportion_of_intersecting_edges_over_edges_threshold}      
\end{align}


\subsection{A generalization}

$\E[P_c^t]$, $\E[Q]$, $\E[P_c^e]$ are particular cases of a general definition of an evaluation score
\begin{equation}
K = c \E[\phi(n)],
\label{eq:generic_K_score}
\end{equation}
where $\E[\phi(n)]$ is defined as in Eq. \ref{eq:generic_expectation}.
$\E[P_c^t]$ is obtained with 
\begin{align*}
c       & = 1 \\
\phi(n) & = n^{2 - n}.
\end{align*}
$\E[Q]$ is obtained with 
\begin{align*}
c       & = 2 \\
\phi(n) & = \frac{1}{n}.
\end{align*}
$\E[P_c^e]$ is obtained with
\begin{align*}
c       & = \frac{2S(n_{min})}{m(n_{min})} \\
\phi(n) & = 1 - \frac{1}{n}.
\end{align*}

\subsection{The calculation of the theoretical performance of the random parser}

For simplicity, we calculate $\E[P_c^e]$ from $\E[Q]$ applying Eq. \ref{eq:relationship_between_expected_performance}.

To calculate $\E[P_c^t]$ and $\E[Q]$, we apply the definition of $K$ (Eq. \ref{eq:generic_K_score}). $K$ is easy to compute numerically if the distribution of sequences lengths is the uniform or the empirical distribution because the summation in the definition of $K$ (Eq. \ref{eq:generic_expectation}) has a finite number of summands. It suffices to plug the definition of the uniform distribution (Eq. \ref{eq:uniform_distribution}) or that of the empirical distribution (Eq. \ref{eq:empirical_sequence_length_probability}) into 
Eq. \ref{eq:generic_expectation}.

If the distribution is the geometric, the number of summands is infinite, and then  two approaches are possible. First, a high-quality approximation based on $K$ that is explained in Appendix \ref{app:calculation_of_the_expected_performance_of_random_parser}. The method has a parameter $\epsilon$ that is the approximation error. In this article, we use $\epsilon = 10^{-8}$.
Second, a specific analytical solution for $\E[Q] = 2 \E[1/n]$ that consists of replacing $\E[1/n]$ by an exact formula in Property \ref{prop:expectations_for_geometric_distribution}. 
As a complement, Property \ref{prop:expectations_for_uniform_distribution} also gives an exact formula for the uniform distribution. 

\subsection{The theoretical performance of the random parser}
\label{subsec:theoretical_performance_of_random_parser}

\begin{figure}
    \centering
    \includegraphics[width=0.9\linewidth]{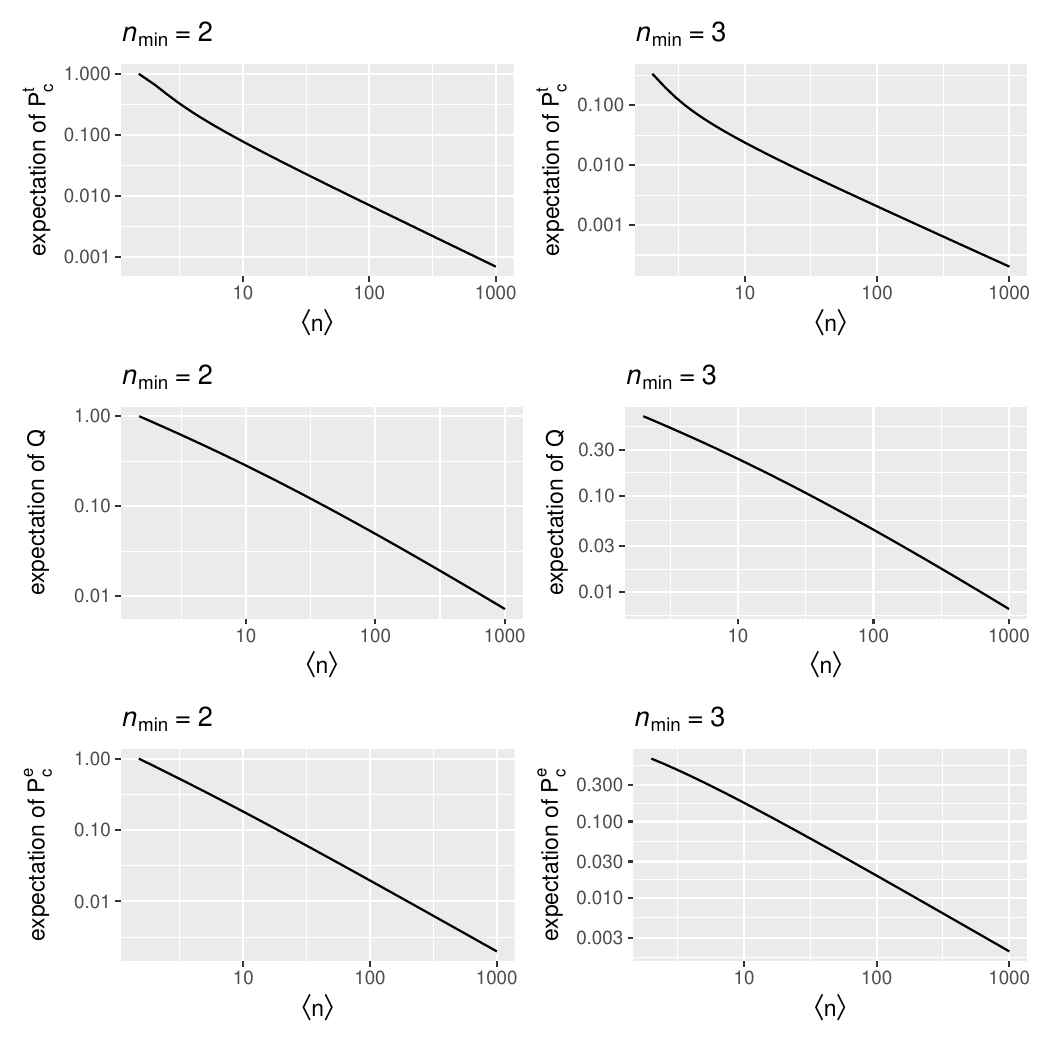}
    \caption{The performance of a random parser as a function of the expected sequence length $\left<n \right>$ when sequence length follows a uniform  distribution in the interval $[1, n_{max}]$ (Eq. \ref{eq:uniform_distribution}).     
    Three performance scores are considered: $\E[P_c^t]$, the expected proportion of correct trees (top), $\E[Q]$, the average expected proportion of correct edges per sequence (middle) and $\E[P_c^e]$, the expected overall proportion of correct edges (bottom). On top each subfigure, $n_{min}$ indicates the minimum sequence length considered to measure the performance of the parser.
    \label{fig:performance_uniform_distribution_mean_sequence_length_as_predictor}  
    }
\end{figure}

\begin{figure}
    \centering
    \includegraphics[width=0.9\linewidth]{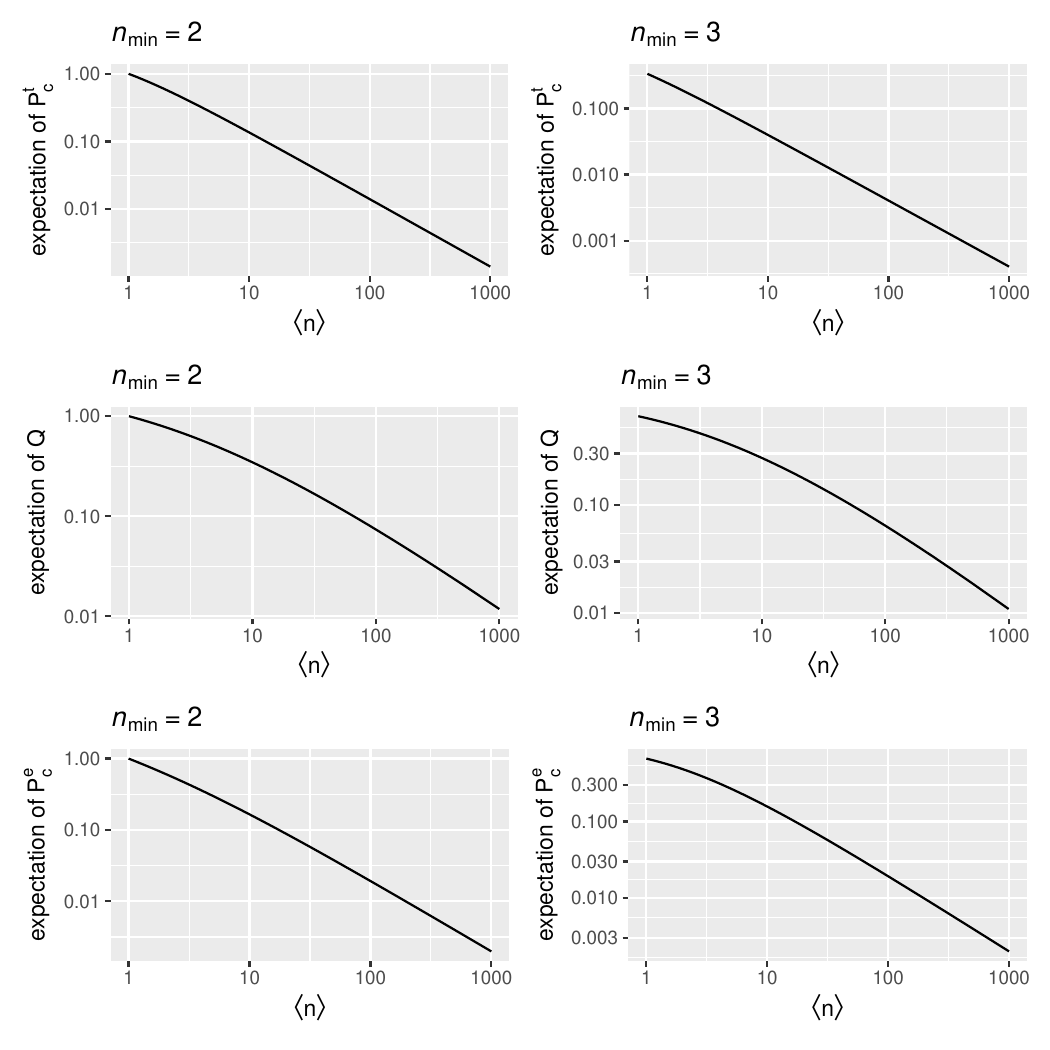}
    \caption{The performance of a random parser as a function of the expected sequence length $\left<n \right>$ when sequence length follows a 1-parameter geometric distribution (Eq. \ref{eq:plain_geometric_distribution}). The format is the same as in Fig. \ref{fig:performance_uniform_distribution_mean_sequence_length_as_predictor}.     
    }
    \label{fig:performance_geometric_distribution_mean_sequence_length_as_predictor}
\end{figure}

We wish to explore the performance of the random parser on the theoretical distributions. 
To find a common ground for the two distributions, we use $\left<n \right>$, the expected sequence length. For the uniform distribution (Eq. \ref{eq:uniform_distribution}),   
\begin{equation*}
\left<n \right> = (1 + n_{max})/2.    
\end{equation*}
For the 1-parameter geometric distribution (Eq. \ref{eq:plain_uniform_distribution}), $\left<n \right> = 1/q$.
The performance decreases as the mean sequence length increases for both distributions (Fig. \ref{fig:performance_uniform_distribution_mean_sequence_length_as_predictor} and 
\ref{fig:performance_geometric_distribution_mean_sequence_length_as_predictor}). For sufficiently long sequences, the decay is power-law like given the appearance of a straight line in double logarithmic scale. That is 
\begin{equation*}
\E[P_c^t], \E[Q], \E[P_c^e] \sim \left<n\right>^{-\alpha},    
\end{equation*}
where $\alpha$ is some positive constant. 
See Appendix \ref{app:theoretical_performance_of_random_parser} for the performance of the random parser as direct function of the parameter of the uniform and the geometric distribution.

\section{Methods}

\label{sec:methods}

\subsection{Non-human primates sequences}

We borrow the maximum sequence length information on 31 species of primates (Table \ref{tab:max_sequence_length_in_primates}) from \citet{Girard-Buttoz2022a}. 
We borrow the empirical sequence length distribution from previous work: 
\begin{itemize}
\item
Vocal sequences produced by male geladas from \citet[Figure S2]{Gustison2016a}. 
\item 
Chimpanzee vocal sequences from the data displayed in Figure 1 of \citet{Girard-Buttoz2022a} That data is available as Supplementary Data 2 from \citet{Girard-Buttoz2022a}. 
For simplicity, we consider all calls, independently of the sex of the vocalizer and we combine the counts for unpanted calls (calls emitted individually) and counts for panted calls (calls interspersed with voiced inhalations) to obtain a single count for sequences of length 1. 
\item 
Chimpanzee gestural sequences from the counts displayed in Table 1 of \cite{Mielke2024a}. Each gesture action has a clearly defined Minimum Action Unit (MAU) based on the minimum information necessary to distinguish between gesture actions, starting from the moment the individual moves the body part (or articulator) and finishing when the gesture action is fully in place \cite{Grund2023a}.
Unlike vocalizations, which are restricted to being produced one at a time, two or more gesture tokens can be overlapped.  Thus \citet{Mielke2024a} employed four criteria to define a sequence (from broader to more restrictive): 5 seconds (gestures that started within 5 seconds after the end of the MAU of the previous unit were considered to be in the same sequence), rapid-fire sequences (gestures that started within 1 second after the end of the MAU of the previous
unit considered to be in the same sequence), overlap (only gestures that occurred with overlap of their MAU considered to be in the same sequence) and solitary gestures-plus waiting (gestures that were separated by at least one second, but occurred within the same 5 second window considered to be in the same sequence).
\end{itemize}
The statistical properties of these distributions are summarized in Table \ref{tab:summary}.

\begin{landscape}

\begin{table}[]
    \centering
    \caption{\label{tab:summary} For each kind of sequence (modality and criterion), we show the sequence length ($S$), the total number of tokens ($T$), the maximum sequence length ($l_{max}$), the average sequence length ($\left<n \right>$) and the frequency of sequences of specific length. $f(n)$ is the number of sequences of length $n$ and $f_\geq(n)$ is the number of sequences of length $n$ or greater. }
    \begin{tabular}{lllrrrrrrrr}
    species & modality & criterion & $S$ & $T$ & $l_{max}$ & $\left<n \right>$ & $f(1)$ & $f(2)$ & $f(3)$ & $f_\geq (4)$ \\
    \hline
    gelada & vocal & none & 1065 & 4747 & 26 & 4.46 & 242 & 138 & 156 & 126\\
chimpanzee & vocal & none & 4826 & 7708 & 10 & 1.6 & 3242 & 817 & 458 & 170\\
chimpanzee & gestural & 5 seconds & 5161 & 7747 & 13 & 1.5 & 3616 & 1006 & 302 & 109\\
chimpanzee & gestural & rapid-fire sequences & 6356 & 7747 & 8 & 1.22 & 5343 & 772 & 152 & 56\\
chimpanzee & gestural & overlap & 6954 & 7747 & 6 & 1.11 & 6279 & 589 & 64 & 13\\
chimpanzee & gestural & solitary gestures & 6966 & 7747 & 10 & 1.11 & 6406 & 420 & 97 & 23\\

    \end{tabular}
\end{table}

\end{landscape}

\begin{table}
\caption{\label{tab:max_sequence_length_in_primates} The 31 species of primates sorted by $n_{max}$ the maximum sequence length. Extracted from \citet[Supplementary Data 2]{Girard-Buttoz2022a}. In the taxa field, A-E stands for Afro-Eurasian and A for American (Afro-Eurasian and American are modern replacements for the labels Old World and New World). In the `Singing' column, 'yes' indicates a singing species while 'no' indicates a non-singing species following \citet{Girard-Buttoz2022a}. 
}
{\small
\begin{tabular}{rllll}
$n_{max}$ &	Species	&	Scientific name	&	Taxa	&	Singing \\
\hline
2	&	Sahamalaza sportive lemur	&	Lepilemur sahamalazensis	&	Prosimian	&	no	\\
2	&	Mongoose lemur	&	Eulemur mongoz	&	Prosimian	&	no	\\
2	&	Crowned lemur	&	Eulemur coronatus	&	Prosimian	&	no	\\
2	&	Thomas langur	&	Presbytis thomasi	&	A-E monkey	&	no	\\
2	&	Sooty mangabey	&	Cercocebus torquatus atys	&	A-E monkey	&	no	\\
2	&	Red-capped mangabeys	&	Cercocebus torquatus	&	A-E monkey	&	no	\\
2	&	Putty-nosed monkey	&	Cercopithecus nictitans	&	A-E monkey	&	no	\\
2	&	Olive baboon	&	Papio anubis	&	A-E monkey	&	no	\\
2	&	Diana monkey	&	Cercopithecus diana	&	A-E monkey	&	no	\\
2	&	DeBrazza's monkeys	&	Cercopithecus neglectus	&	A-E monkey	&	no	\\
2	&	Chacma baboon	&	Papio ursinus	&	A-E monkey	&	no	\\
2	&	Campbell's monkey	&	Cercopithecus campbelli	&	A-E monkey	&	no	\\
2	&	Blue monkey	&	Cercopithecus mitis stulmanni	&	A-E monkey	&	no	\\
3	&	Red-bellied lemur	&	Eulemur rubriventer	&	Prosimian	&	no	\\
3	&	Northern giant mouse lemur	&	Mirza mirza	&	Prosimian	&	no	\\
3	&	Indri	&	Indri indri	&	Prosimian	&	yes	\\
3	&	Common brown lemur	&	Eulemur fulvus	&	Prosimian	&	no	\\
3	&	Silvery marmoset	&	Mico argentatus	&	A monkey	&	no	\\
3	&	Pygmy marmoset	&	Cebuella pygmaea	&	A monkey	&	no	\\
3	&	Goeldi's marmoset	&	Callimico goeldii	&	A monkey	&	no	\\
3	&	Common marmoset	&	Callithrix jacchus	&	A monkey	&	no	\\
3	&	Olive colobus monkey	&	Procolobus verus	&	A-E monkey	&	no	\\
3	&	Orangutan	&	Pongo pygmaeus 	&	Ape	&	no	\\
3	&	Gorilla	&	Gorilla gorilla	&	Ape	&	no	\\
3	&	Bonobo	&	Pan paniscus	&	Ape	&	no	\\
4	&	Golden lion tamarin	&	Leontopithecus rosalia	&	A monkey	&	no	\\
5	&	Philippine tarsier	&	Tarsius syrichta fraterculus	&	Prosimian	&	yes	\\
5	&	White-handed/Lar gibbon	&	Hylobates Lar	&	Ape	&	yes	\\
6	&	Gelada	&	Theropithecus gelada	&	A-E monkey	&	no	\\
8	&	Chimpanzee	&	Pan troglodytes	&	Ape	&	no	\\
13	&	Agile gibbon	&	Hylobates agilis	&	Ape	&	yes	\\
\end{tabular}
}
\end{table}

\subsection{Human language sequences}

The standard for the evaluation of unsupervised parsing in computational linguistics is WSJ10, a subset of the Wall Street Journal (WSJ) corpus comprising sentences of at most 10 non-punctuation words \cite{Han2020a,Soegaard2011a}. 
However, the language of WSJ is English which is a single a WEIRD language from the Indo-European family \cite{Blasi2022a}. 
To reach more generalizable conclusions, we need a sample of  languages from distinct families that is fully parallel so that we do not need to control for the content of the texts and other characteristics. For this reason, we use Parallel Universal Dependencies (PUD), a parallel collection of syntactic dependency treebanks that is a subset of the Universal Dependencies (UD) collection \cite{ud26_APS}. We extract PUD from UD 2.18, which leads to a sample of languages 21 languages from 9 linguistic families (Table \ref{tab:PUD_languages}). Following the analogy of WSJ10, we define PUD10 as the subset of PUD comprising sentences of at most 10 non-punctuation words. 
To control for the effect of syntactic annotation style, we consider the UD style, i.e. the original annotation style of the UD collection, as well as a competing style, i.e. Surface-syntactic Universal Dependencies (SUD) \cite{sud}.

We borrow the preprocessing methods from previous research on parallel dependency treebanks \cite{Ferrer2020b, Alemany2022c}. The main features of the processing are that nodes that are punctuation marks are removed and that the corpus remains fully parallel after the removal \cite{Ferrer2020b}. The preprocessed data is freely available the ancillary materials of the Linear Arrangement Library website at \url{https://cqllab.upc.edu/lal/universal-dependencies/} as PUD 2.18 for UD style and PSUD 2.18 for SUD style. In these preprocessed treebanks, the minimum sequence length in is $n = 2$ because the syntactic dependency structure is empty for $n < 2$ and sentences with $n < 2$ are discarded.
\iftoggle{PSUD}{}
{After preprocessing, the empirical length distribution in UD style is the same as that of SUD style for each language. 
For this reason, we discard SUD style.
}

\begin{table}
\caption{\label{tab:PUD_languages} The languages in the PUD collection grouped by linguistic family. }
\begin{tabularx}{\textwidth}{lXX}
Family & Languages \\
\hline
Afro-Asiatic & Arabic \\
Austronesian & Indonesian \\
Koreanic & Korean \\
Indo-European & Czech, English, French, Galician, German, Hindi, Icelandic, Italian, Polish, Portuguese, Russian, Spanish, Swedish \\
Japonic & Japanese \\
Sino-Tibetan & Chinese \\ 
Tai-Kadai & Thai \\
Turkic & Turkish \\
Uralic & Finnish \\
\end{tabularx}
\end{table}

\section{Results}

\label{sec:results}

\subsection{Empirical sequence length distribution}

\label{subsec:empirical_sequence_length_distribution}

We examine $p(n)$, the proportion of sequence of length $n$. The goal is to further motivate the theoretical assumptions about $p(n)$ that we follow in subsequent analyses. Finding the best model for $p(n)$ is beyond the scope of the current article.

\begin{figure}
    \centering
    \includegraphics[width=0.9\linewidth]{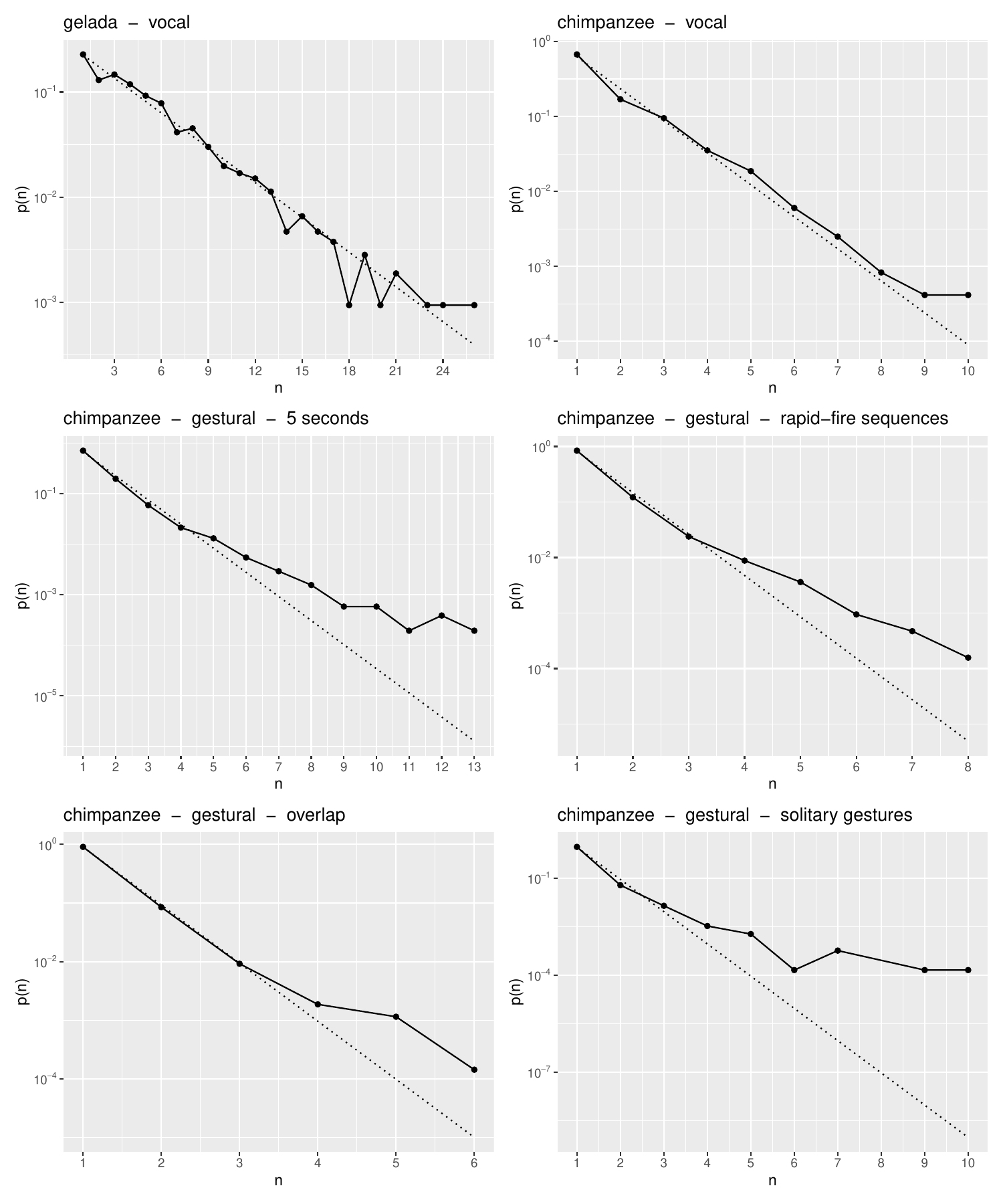}
    \caption{\label{fig:sequence_length_distribution} The empirical distribution of sequence lengths (solid line) in sequences produced by geladas and chimpanzees. On top each subfigure, the name of the species, the modality and the criterion that defines the sequences is indicated. The dotted line shows the best fit of a 2-parameter geometric distribution (Eq. \ref{eq:geometric_distribution}) with parameters $n_{min}=1$ and 
    $q = \frac{(S - 1)/S}{\left<n\right> - 1/S}$ where $S$ is the number of sequences and $\left<n\right>$ is the average sequence length
    (Appendix \ref{app:probability_distributions_and_expectations}). }
    
\end{figure}

In chimpanzees, $p(n)$ shows a linear decay when logarithmic scale is applied to $p(n)$ that is consistent with an exponential function (Figure \ref{fig:sequence_length_distribution}). 
In vocal sequences, a 1-parameter geometric distribution (Eq. \ref{eq:plain_geometric_distribution}) gives a visual fit of high quality for the whole range of lengths for geladas and chimpanzees. In gestural sequences (chimpanzees), a 1-parameter geometric distribution fits only small lengths up to a breakpoint, that is the start of another exponential regime with a slower decay. That is the hallmark of the two-regime exponential distribution that has been found in the distance between syntactically related words in sentences \cite{Petrini2022c} and also in the distance traversed by foraging ants \cite{Campos2016a}.
The relevant conclusions of this qualitative analysis are that, in chimpanzees, 
\begin{enumerate}
\item
Sequence length distributions exhibit a non-increasing trend. The mode is $n = 1$ and $p(n)$ tends to decrease or remain the same as $n$ increases. 
\item 
The 1-parameter geometric distribution yields a good approximation in vocal sequences and a lower bound to the actual decay in gestural sequences. 
\end{enumerate}
These conclusions are in contrast with the distribution of sequence lengths in human languages, where 
\begin{enumerate}
\item
Sequence length distributions exhibit an inverted-V shape (an initial increasing trend is followed by a decreasing trend). The mode is far from $n = 1$ \cite{Sigurd2004a,Furuhashi2012a}.  
\item 
The geometric distribution yields a poor approximation since it is incompatible with an inverted-V shape. The log-normal distribution is often selected as an approximation to the distribution of sentence lengths \cite{Furuhashi2012a}.  
\end{enumerate}
These characteristics are also found in PUD (Appendix \ref{app:sentence_length_distribution}).

\subsection{Performance on primate sequences}

Table \ref{tab:max_sequence_length_in_primates} shows that, among the 31 species of primates reviewed by \citet{Girard-Buttoz2022a}, the number of species for each value of $n_{max}$ is as follows: 2 (13 species), 3 (12 species), 4 (1 species), 5 (2 species), 8 (1 species) and 13 (1 species).

We do not know the actual distribution of sequence lengths in the majority of these species, but if we assume that the distribution is non-increasing (recall Section \ref{subsec:empirical_sequence_length_distribution} and Figure \ref{fig:sequence_length_distribution}), we can then use a uniform distribution to get a lower bound of the performance of a random parser (Table \ref{tab:performance_given_max_sequence_length}). When $n_{max} \leq 2$ the performance of the random parser is trivially $100\%$ ($\E[P_c^t] = \E[Q] = \E[P_c^e] = 1$). When $n_{max} = 3$, the proportion of correctly guessed edges is about $80\%$ if the performance is evaluated on sequences of length $2$ or greater ($\E[Q] = 0.83$ and $\E[P_c^e] = 0.77$); the proportion of correctly guessed edges is nearly $70\%$ if the performance is evaluated on sequences of length $3$ or greater ($\E[Q] = \E[P_c^e] = 0.67$). 

\begin{table}[]
    \centering
    \caption{\label{tab:performance_given_max_sequence_length} A lower bound to the performance of the random parser assuming that the distribution of sequence lengths is non-increasing. $n_{min}$ is the minimum sequence length considered when measuring the performance of the random parser. $n_{max}$ is the maximum sequence length. Only the values of $n_{max}$ reported by \citet{Girard-Buttoz2022a} on an ensemble of 31 primate species are shown (Table \ref{tab:max_sequence_length_in_primates}).}    
    \begin{tabular}{lllll}
     $n_{min}$ & $n_{max}$ & $\E[P_c^t]$ & $\E[Q]$ & $\E[P_c^e]$ \\
     \hline
     2 & 2 & 1 & 1 & 1\\
2 & 3 & 0.667 & 0.833 & 0.778\\
2 & 4 & 0.465 & 0.722 & 0.639\\
2 & 5 & 0.351 & 0.642 & 0.543\\
2 & 8 & 0.201 & 0.491 & 0.377\\
2 & 13 & 0.117 & 0.363 & 0.252\\
3 & 3 & 0.333 & 0.667 & 0.667\\
3 & 4 & 0.198 & 0.583 & 0.567\\
3 & 5 & 0.135 & 0.522 & 0.493\\
3 & 8 & 0.067 & 0.406 & 0.354\\
3 & 13 & 0.037 & 0.305 & 0.242\\

    \end{tabular}    
\end{table}

Now we examine the performance of the random parser when further information about the distribution of sequence lengths is exploited in humans, geladas and chimpanzees.
In all our analyses, we find \iftoggle{PSUD}{
(Tables \ref{tab:performance_by_distribution}, \ref{tab:performance_by_distribution_PUD10_UD}, \ref{tab:performance_by_distribution_PUD10_SUD},  \ref{tab:performance_by_distribution_PUD_UD} and \ref{tab:performance_by_distribution_PUD_SUD})
} 
{
(Table \ref{tab:performance_by_distribution}, Table \ref{tab:performance_by_distribution_PUD10_UD} and Table \ref{tab:performance_by_distribution_PUD_UD})
}
\begin{enumerate}
    \item 
    $\E[Q] \geq \E[P_c^t]$ as theoretically expected (Section \ref{sec:theory}). 
    \item 
$\E[Q] \geq \E[P_c^e]$ for the empirical distribution 
because the actual $\E[Q]$ is large enough, namely $\E[Q]_*$ (Eq. \ref{eq:expected_proportion_of_intersecting_edges_over_edges_threshold}) never exceeds $\E[Q]$.    
\end{enumerate}

In geladas and chimpanzees, the performance of the random parser drops as the minimum sentence length for evaluation increases from 2 to 3 (Tables \ref{tab:performance_by_score} and \ref{tab:performance_by_distribution}). The values of $\E[P_c^t]$ are lower than the corresponding values of the other scores ($\E[Q]$ as expected theoretically (Eq. \ref{eq:guessing_edges_is_easier_than_guessing_trees}) but also those of $\E[P_c^e]$). These findings are consistent with the intuition that guessing the right tree is more difficult than guessing the right edges (Table \ref{tab:performance_by_score}).
As expected from the geometric-like distribution of sequence lengths in primates (Section \ref{subsec:empirical_sequence_length_distribution}), the performance of the random parser assuming a 1-parameter geometric distribution for sentence lengths yields a better estimate of the actual performance of the parser than a uniform distribution (Table 
\ref{tab:performance_by_distribution}). 
The average proportion of correct edges per sequence ($\E[Q]$) in sequences of two or more units is $51\%$ for geladas, $79\%$ for chimpanzee vocalizations and $>84\%$ for chimpanzee gestures. Setting $n_{min} \geq 3$ to exclude sequences where no error is possible, the average proportion drops to $41\%$ for geladas, $57\%$ for chimpanzee vocalizations and $>56\%$ for chimpanzee gestures.  

The performance of the random parser on human languages on sentences of up to 10 words is much worse than in geladas and chimpanzees \iftoggle{PSUD}{(see Table \ref{tab:performance_by_score_PUD10_UD} and Table \ref{tab:performance_by_distribution_PUD10_UD} for UD annotation style; see Appendix \ref{app:performance_of_random_parser_PUD10_SUD} for SUD annotation style).}{
(Table \ref{tab:performance_by_score_PUD10_UD} and Table \ref{tab:performance_by_distribution_PUD10_UD}).
}
The expected proportion of correct edges per sequence ($\E[Q]$) in sequences of two or more units in languages is $28\%$ on average for $n_{min} = 2$ and also for $n_{min} = 3$.   
Not surprisingly, the performance of the random parser on human languages is even worse if no limit is imposed on sentence length (Appendix \ref{app:performance_of_random_parser_without_sentence_length_limit}).
The expected proportion of correct edges per sequence in languages drops to  $13\%$ on average for $n_{min} = 2$ and also for $n_{min} = 3$.   

\begin{landscape}

\begin{table}
    \centering
    \caption{\label{tab:performance_by_score} The performance scores $\E[P_c^t]$, $\E[Q]$, $\E[P_c^e]$ on geladas and chimpanzees. Results are sorted by minimum sequence length for evaluation ($n_{min}$), modality, criterion and evaluation score. For each evaluation score, the sequence length distributions considered are a uniform distribution (u), a $1$-parameter geometric distribution (g) and the empirical distribution (e).
    }  
    {\footnotesize
    \begin{tabular}{rlllrrrrrrrrrr}
     & & & & \multicolumn{3}{c}{$\E[P_c^t]$} & \multicolumn{3}{c}{$\E[Q]$} & \multicolumn{3}{c}{$\E[P_c^e]$} \\ \cmidrule(lr){5-7} \cmidrule(lr){8-10} \cmidrule(lr){11-13}
    $n_{min}$ & species & modality & criterion & u & g & e & u & g & e & u & g & e \\
    \hline
    2 & gelada & vocal & none & 0.056 & 0.292 & 0.241 & 0.228 & 0.536 & 0.509 & 0.136 & 0.328 & 0.333\\
2 & chimpanzee & vocal & none & 0.156 & 0.71 & 0.619 & 0.429 & 0.845 & 0.794 & 0.314 & 0.723 & 0.663\\
2 & chimpanzee & gestural & 5 seconds & 0.117 & 0.745 & 0.721 & 0.363 & 0.865 & 0.845 & 0.252 & 0.756 & 0.69\\
2 & chimpanzee & gestural & rapid-fire sequences & 0.201 & 0.871 & 0.816 & 0.491 & 0.934 & 0.902 & 0.377 & 0.874 & 0.8\\
2 & chimpanzee & gestural & overlap & 0.281 & 0.929 & 0.905 & 0.58 & 0.964 & 0.951 & 0.473 & 0.93 & 0.893\\
2 & chimpanzee & gestural & solitary gestures & 0.156 & 0.93 & 0.81 & 0.429 & 0.965 & 0.899 & 0.314 & 0.931 & 0.79\\
3 & gelada & vocal & none & 0.017 & 0.087 & 0.089 & 0.196 & 0.402 & 0.411 & 0.134 & 0.293 & 0.307\\
3 & chimpanzee & vocal & none & 0.051 & 0.224 & 0.214 & 0.357 & 0.586 & 0.575 & 0.299 & 0.545 & 0.529\\
3 & chimpanzee & gestural & 5 seconds & 0.037 & 0.237 & 0.2 & 0.305 & 0.597 & 0.557 & 0.242 & 0.561 & 0.492\\
3 & chimpanzee & gestural & rapid-fire sequences & 0.067 & 0.283 & 0.226 & 0.406 & 0.633 & 0.588 & 0.354 & 0.616 & 0.55\\
3 & chimpanzee & gestural & overlap & 0.101 & 0.305 & 0.258 & 0.475 & 0.648 & 0.613 & 0.436 & 0.639 & 0.585\\
3 & chimpanzee & gestural & solitary gestures & 0.051 & 0.305 & 0.242 & 0.357 & 0.649 & 0.595 & 0.299 & 0.64 & 0.545\\

    \end{tabular}
    }    
\end{table}

\begin{table}
    \centering
    \caption{\label{tab:performance_by_distribution} The performance scores $\E[P_c^t]$, $\E[Q]$, $\E[P_c^e]$ on geladas and chimpanzees. Results are sorted by minimum sequence length for evaluation ($n_{min}$), modality, criterion and distribution of sequence lengths i.e. uniform distribution (u), 1-parameter geometric distribution (g) and the empirical distribution (e).    
    }    
    {\footnotesize
    \begin{tabular}{rlllrrrrrrrrrr}
     & & & & \multicolumn{3}{c}{uniform} & \multicolumn{3}{c}{geometric} & \multicolumn{3}{c}{empirical} \\\cmidrule(lr){5-7} \cmidrule(lr){8-10} \cmidrule(lr){11-13}
    $n_{min}$ & species & modality & criterion & 
    $\E[P_c^t]$ & $\E[Q]$ & $\E[P_c^e]$ &   
    $\E[P_c^t]$ & $\E[Q]$ & $\E[P_c^e]$ &
    $\E[P_c^t]$ & $\E[Q]$ & $\E[P_c^e]$ \\
    \hline
    2 & gelada & vocal & none & 0.056 & 0.228 & 0.136 & 0.292 & 0.536 & 0.328 & 0.241 & 0.509 & 0.333\\
2 & chimpanzee & vocal & none & 0.156 & 0.429 & 0.314 & 0.71 & 0.845 & 0.723 & 0.619 & 0.794 & 0.663\\
2 & chimpanzee & gestural & 5 seconds & 0.117 & 0.363 & 0.252 & 0.745 & 0.865 & 0.756 & 0.721 & 0.845 & 0.69\\
2 & chimpanzee & gestural & rapid-fire sequences & 0.201 & 0.491 & 0.377 & 0.871 & 0.934 & 0.874 & 0.816 & 0.902 & 0.8\\
2 & chimpanzee & gestural & overlap & 0.281 & 0.58 & 0.473 & 0.929 & 0.964 & 0.93 & 0.905 & 0.951 & 0.893\\
2 & chimpanzee & gestural & solitary gestures & 0.156 & 0.429 & 0.314 & 0.93 & 0.965 & 0.931 & 0.81 & 0.899 & 0.79\\
3 & gelada & vocal & none & 0.017 & 0.196 & 0.134 & 0.087 & 0.402 & 0.293 & 0.089 & 0.411 & 0.307\\
3 & chimpanzee & vocal & none & 0.051 & 0.357 & 0.299 & 0.224 & 0.586 & 0.545 & 0.214 & 0.575 & 0.529\\
3 & chimpanzee & gestural & 5 seconds & 0.037 & 0.305 & 0.242 & 0.237 & 0.597 & 0.561 & 0.2 & 0.557 & 0.492\\
3 & chimpanzee & gestural & rapid-fire sequences & 0.067 & 0.406 & 0.354 & 0.283 & 0.633 & 0.616 & 0.226 & 0.588 & 0.55\\
3 & chimpanzee & gestural & overlap & 0.101 & 0.475 & 0.436 & 0.305 & 0.648 & 0.639 & 0.258 & 0.613 & 0.585\\
3 & chimpanzee & gestural & solitary gestures & 0.051 & 0.357 & 0.299 & 0.305 & 0.649 & 0.64 & 0.242 & 0.595 & 0.545\\

    \end{tabular}
    }
\end{table}

\end{landscape}

\begin{table}
    \centering
    \caption{\label{tab:performance_by_score_PUD10_UD} The performance scores $\E[P_c^t]$, $\E[Q]$, $\E[P_c^e]$ on human languages in the PUD10 collection (sentences of up to length 10 in PUD). Results are sorted by minimum sequence length for evaluation ($n_{min}$), language and evaluation score. For each evaluation score, the sequence length distributions considered are a uniform distribution (u), a $1$-parameter geometric distribution (g) and the empirical distribution (e). 
    }    
    {\footnotesize
    \begin{tabular}{rlrrrrrrrrr}
     & & \multicolumn{3}{c}{$\E[P_c^t]$} & \multicolumn{3}{c}{$\E[Q]$} & \multicolumn{3}{c}{$\E[P_c^e]$} \\\cmidrule(lr){3-5} \cmidrule(lr){6-8} \cmidrule(lr){9-11}
    $n_{min}$ & language & u & g & e & u & g & e & u & g & e \\
    \hline
    2 & Arabic & 0.156 & 0.198 & 0.03 & 0.429 & 0.433 & 0.292 & 0.314 & 0.233 & 0.255\\
2 & Chinese & 0.156 & 0.197 & 0.019 & 0.429 & 0.431 & 0.284 & 0.314 & 0.231 & 0.255\\
2 & Czech & 0.156 & 0.198 & 0.019 & 0.429 & 0.432 & 0.286 & 0.314 & 0.233 & 0.255\\
2 & English & 0.156 & 0.227 & 0.009 & 0.429 & 0.467 & 0.275 & 0.314 & 0.263 & 0.254\\
2 & Finnish & 0.156 & 0.202 & 0.016 & 0.429 & 0.437 & 0.289 & 0.314 & 0.237 & 0.26\\
2 & French & 0.156 & 0.223 & 0.012 & 0.429 & 0.463 & 0.276 & 0.314 & 0.259 & 0.251\\
2 & Galician & 0.156 & 0.203 & 0.036 & 0.429 & 0.439 & 0.303 & 0.314 & 0.238 & 0.261\\
2 & German & 0.156 & 0.223 & 0.007 & 0.429 & 0.463 & 0.273 & 0.314 & 0.259 & 0.25\\
2 & Hindi & 0.156 & 0.223 & 0.01 & 0.429 & 0.462 & 0.272 & 0.314 & 0.258 & 0.25\\
2 & Icelandic & 0.156 & 0.223 & 0.01 & 0.429 & 0.463 & 0.274 & 0.314 & 0.259 & 0.25\\
2 & Indonesian & 0.156 & 0.194 & 0.018 & 0.429 & 0.427 & 0.278 & 0.314 & 0.228 & 0.251\\
2 & Italian & 0.156 & 0.198 & 0.018 & 0.429 & 0.432 & 0.286 & 0.314 & 0.232 & 0.256\\
2 & Japanese & 0.156 & 0.206 & 0.009 & 0.429 & 0.443 & 0.257 & 0.314 & 0.241 & 0.238\\
2 & Korean & 0.156 & 0.194 & 0.014 & 0.429 & 0.427 & 0.277 & 0.314 & 0.228 & 0.251\\
2 & Polish & 0.156 & 0.193 & 0.023 & 0.429 & 0.426 & 0.284 & 0.314 & 0.227 & 0.249\\
2 & Portuguese & 0.156 & 0.193 & 0.023 & 0.429 & 0.426 & 0.284 & 0.314 & 0.227 & 0.249\\
2 & Russian & 0.156 & 0.195 & 0.023 & 0.429 & 0.429 & 0.283 & 0.314 & 0.23 & 0.252\\
2 & Spanish & 0.156 & 0.2 & 0.032 & 0.429 & 0.434 & 0.295 & 0.314 & 0.234 & 0.257\\
2 & Swedish & 0.156 & 0.195 & 0.013 & 0.429 & 0.428 & 0.277 & 0.314 & 0.229 & 0.252\\
2 & Thai & 0.156 & 0.192 & 0.017 & 0.429 & 0.425 & 0.279 & 0.314 & 0.226 & 0.25\\
2 & Turkish & 0.156 & 0.2 & 0.027 & 0.429 & 0.434 & 0.292 & 0.314 & 0.234 & 0.256\\
3 & Arabic & 0.051 & 0.058 & 0.01 & 0.357 & 0.334 & 0.277 & 0.299 & 0.216 & 0.253\\
3 & Chinese & 0.051 & 0.058 & 0.012 & 0.357 & 0.332 & 0.278 & 0.299 & 0.214 & 0.254\\
3 & Czech & 0.051 & 0.058 & 0.01 & 0.357 & 0.333 & 0.279 & 0.299 & 0.215 & 0.254\\
3 & English & 0.051 & 0.067 & 0.009 & 0.357 & 0.357 & 0.275 & 0.299 & 0.24 & 0.254\\
3 & Finnish & 0.051 & 0.06 & 0.013 & 0.357 & 0.337 & 0.287 & 0.299 & 0.219 & 0.26\\
3 & French & 0.051 & 0.066 & 0.012 & 0.357 & 0.354 & 0.276 & 0.299 & 0.237 & 0.251\\
3 & Galician & 0.051 & 0.06 & 0.016 & 0.357 & 0.338 & 0.289 & 0.299 & 0.22 & 0.259\\
3 & German & 0.051 & 0.066 & 0.007 & 0.357 & 0.354 & 0.273 & 0.299 & 0.237 & 0.25\\
3 & Hindi & 0.051 & 0.066 & 0.01 & 0.357 & 0.353 & 0.272 & 0.299 & 0.237 & 0.25\\
3 & Icelandic & 0.051 & 0.066 & 0.01 & 0.357 & 0.354 & 0.274 & 0.299 & 0.237 & 0.25\\
3 & Indonesian & 0.051 & 0.057 & 0.012 & 0.357 & 0.33 & 0.274 & 0.299 & 0.211 & 0.25\\
3 & Italian & 0.051 & 0.058 & 0.009 & 0.357 & 0.333 & 0.279 & 0.299 & 0.215 & 0.255\\
3 & Japanese & 0.051 & 0.061 & 0.009 & 0.357 & 0.34 & 0.257 & 0.299 & 0.223 & 0.238\\
3 & Korean & 0.051 & 0.057 & 0.01 & 0.357 & 0.33 & 0.274 & 0.299 & 0.212 & 0.251\\
3 & Polish & 0.051 & 0.057 & 0.014 & 0.357 & 0.329 & 0.277 & 0.299 & 0.211 & 0.248\\
3 & Portuguese & 0.051 & 0.057 & 0.014 & 0.357 & 0.329 & 0.277 & 0.299 & 0.211 & 0.248\\
3 & Russian & 0.051 & 0.057 & 0.013 & 0.357 & 0.331 & 0.276 & 0.299 & 0.213 & 0.251\\
3 & Spanish & 0.051 & 0.059 & 0.011 & 0.357 & 0.335 & 0.28 & 0.299 & 0.217 & 0.255\\
3 & Swedish & 0.051 & 0.057 & 0.007 & 0.357 & 0.331 & 0.273 & 0.299 & 0.212 & 0.252\\
3 & Thai & 0.051 & 0.056 & 0.005 & 0.357 & 0.328 & 0.27 & 0.299 & 0.21 & 0.248\\
3 & Turkish & 0.051 & 0.059 & 0.015 & 0.357 & 0.335 & 0.283 & 0.299 & 0.217 & 0.255\\

    \end{tabular}
    }
\end{table}

\begin{table}
    \centering
\caption{\label{tab:performance_by_distribution_PUD10_UD} The performance scores $\E[P_c^t]$, $\E[Q]$, $\E[P_c^e]$ on human languages in the PUD10 collection (sentences of up to length 10 in PUD). Results are sorted by minimum sequence length for evaluation ($n_{min}$), language and distribution of sequence lengths, i.e. uniform distribution (u), a $1$-parameter geometric distribution (g) and the empirical distribution (e).
    }    
    {\footnotesize
    \begin{tabular}{rlrrrrrrrrr}
     & & \multicolumn{3}{c}{uniform} & \multicolumn{3}{c}{geometric} & \multicolumn{3}{c}{empirical} \\\cmidrule(lr){3-5} \cmidrule(lr){6-8} \cmidrule(lr){9-11}
    $n_{min}$ & language & 
    $\E[P_c^t]$ & $\E[Q]$ & $\E[P_c^e]$ &   
    $\E[P_c^t]$ & $\E[Q]$ & $\E[P_c^e]$ &
    $\E[P_c^t]$ & $\E[Q]$ & $\E[P_c^e]$ \\
    \hline
    2 & Arabic & 0.156 & 0.429 & 0.314 & 0.198 & 0.433 & 0.233 & 0.03 & 0.292 & 0.255\\
2 & Chinese & 0.156 & 0.429 & 0.314 & 0.197 & 0.431 & 0.231 & 0.019 & 0.284 & 0.255\\
2 & Czech & 0.156 & 0.429 & 0.314 & 0.198 & 0.432 & 0.233 & 0.019 & 0.286 & 0.255\\
2 & English & 0.156 & 0.429 & 0.314 & 0.227 & 0.467 & 0.263 & 0.009 & 0.275 & 0.254\\
2 & Finnish & 0.156 & 0.429 & 0.314 & 0.202 & 0.437 & 0.237 & 0.016 & 0.289 & 0.26\\
2 & French & 0.156 & 0.429 & 0.314 & 0.223 & 0.463 & 0.259 & 0.012 & 0.276 & 0.251\\
2 & Galician & 0.156 & 0.429 & 0.314 & 0.203 & 0.439 & 0.238 & 0.036 & 0.303 & 0.261\\
2 & German & 0.156 & 0.429 & 0.314 & 0.223 & 0.463 & 0.259 & 0.007 & 0.273 & 0.25\\
2 & Hindi & 0.156 & 0.429 & 0.314 & 0.223 & 0.462 & 0.258 & 0.01 & 0.272 & 0.25\\
2 & Icelandic & 0.156 & 0.429 & 0.314 & 0.223 & 0.463 & 0.259 & 0.01 & 0.274 & 0.25\\
2 & Indonesian & 0.156 & 0.429 & 0.314 & 0.194 & 0.427 & 0.228 & 0.018 & 0.278 & 0.251\\
2 & Italian & 0.156 & 0.429 & 0.314 & 0.198 & 0.432 & 0.232 & 0.018 & 0.286 & 0.256\\
2 & Japanese & 0.156 & 0.429 & 0.314 & 0.206 & 0.443 & 0.241 & 0.009 & 0.257 & 0.238\\
2 & Korean & 0.156 & 0.429 & 0.314 & 0.194 & 0.427 & 0.228 & 0.014 & 0.277 & 0.251\\
2 & Polish & 0.156 & 0.429 & 0.314 & 0.193 & 0.426 & 0.227 & 0.023 & 0.284 & 0.249\\
2 & Portuguese & 0.156 & 0.429 & 0.314 & 0.193 & 0.426 & 0.227 & 0.023 & 0.284 & 0.249\\
2 & Russian & 0.156 & 0.429 & 0.314 & 0.195 & 0.429 & 0.23 & 0.023 & 0.283 & 0.252\\
2 & Spanish & 0.156 & 0.429 & 0.314 & 0.2 & 0.434 & 0.234 & 0.032 & 0.295 & 0.257\\
2 & Swedish & 0.156 & 0.429 & 0.314 & 0.195 & 0.428 & 0.229 & 0.013 & 0.277 & 0.252\\
2 & Thai & 0.156 & 0.429 & 0.314 & 0.192 & 0.425 & 0.226 & 0.017 & 0.279 & 0.25\\
2 & Turkish & 0.156 & 0.429 & 0.314 & 0.2 & 0.434 & 0.234 & 0.027 & 0.292 & 0.256\\
3 & Arabic & 0.051 & 0.357 & 0.299 & 0.058 & 0.334 & 0.216 & 0.01 & 0.277 & 0.253\\
3 & Chinese & 0.051 & 0.357 & 0.299 & 0.058 & 0.332 & 0.214 & 0.012 & 0.278 & 0.254\\
3 & Czech & 0.051 & 0.357 & 0.299 & 0.058 & 0.333 & 0.215 & 0.01 & 0.279 & 0.254\\
3 & English & 0.051 & 0.357 & 0.299 & 0.067 & 0.357 & 0.24 & 0.009 & 0.275 & 0.254\\
3 & Finnish & 0.051 & 0.357 & 0.299 & 0.06 & 0.337 & 0.219 & 0.013 & 0.287 & 0.26\\
3 & French & 0.051 & 0.357 & 0.299 & 0.066 & 0.354 & 0.237 & 0.012 & 0.276 & 0.251\\
3 & Galician & 0.051 & 0.357 & 0.299 & 0.06 & 0.338 & 0.22 & 0.016 & 0.289 & 0.259\\
3 & German & 0.051 & 0.357 & 0.299 & 0.066 & 0.354 & 0.237 & 0.007 & 0.273 & 0.25\\
3 & Hindi & 0.051 & 0.357 & 0.299 & 0.066 & 0.353 & 0.237 & 0.01 & 0.272 & 0.25\\
3 & Icelandic & 0.051 & 0.357 & 0.299 & 0.066 & 0.354 & 0.237 & 0.01 & 0.274 & 0.25\\
3 & Indonesian & 0.051 & 0.357 & 0.299 & 0.057 & 0.33 & 0.211 & 0.012 & 0.274 & 0.25\\
3 & Italian & 0.051 & 0.357 & 0.299 & 0.058 & 0.333 & 0.215 & 0.009 & 0.279 & 0.255\\
3 & Japanese & 0.051 & 0.357 & 0.299 & 0.061 & 0.34 & 0.223 & 0.009 & 0.257 & 0.238\\
3 & Korean & 0.051 & 0.357 & 0.299 & 0.057 & 0.33 & 0.212 & 0.01 & 0.274 & 0.251\\
3 & Polish & 0.051 & 0.357 & 0.299 & 0.057 & 0.329 & 0.211 & 0.014 & 0.277 & 0.248\\
3 & Portuguese & 0.051 & 0.357 & 0.299 & 0.057 & 0.329 & 0.211 & 0.014 & 0.277 & 0.248\\
3 & Russian & 0.051 & 0.357 & 0.299 & 0.057 & 0.331 & 0.213 & 0.013 & 0.276 & 0.251\\
3 & Spanish & 0.051 & 0.357 & 0.299 & 0.059 & 0.335 & 0.217 & 0.011 & 0.28 & 0.255\\
3 & Swedish & 0.051 & 0.357 & 0.299 & 0.057 & 0.331 & 0.212 & 0.007 & 0.273 & 0.252\\
3 & Thai & 0.051 & 0.357 & 0.299 & 0.056 & 0.328 & 0.21 & 0.005 & 0.27 & 0.248\\
3 & Turkish & 0.051 & 0.357 & 0.299 & 0.059 & 0.335 & 0.217 & 0.015 & 0.283 & 0.255\\

    \end{tabular}
    }
\end{table}

\section{Discussion}

\label{sec:discussion}

\subsection{The feasibility of evaluation}

Assuming a uniform or a geometric distribution of sequence lengths, we have demonstrated theoretically that the performance of a random parser increases as the average sequence length decreases (Fig. \ref{fig:performance_uniform_distribution_mean_sequence_length_as_predictor} and Fig. \ref{fig:performance_geometric_distribution_mean_sequence_length_as_predictor}). 

Although guessing the right tree is a hard task, guessing a large fraction of the correct edges is feasible in non-human primates but much harder in humans. In particular, we have shown that the performance of a random parser on non-human primate sequences can be high. That happens when the maximum sequence length is small enough in a large ensemble of primates (Table \ref{tab:performance_given_max_sequence_length}) or when the length distribution decays quickly, as in geladas and chimpanzees (Tables \ref{tab:performance_by_score} and \ref{tab:performance_by_distribution}). 
In contrast, the random parser has a worse performance in humans\iftoggle{PSUD}{
(Table \ref{tab:performance_by_distribution_PUD10_UD}, Table \ref{tab:performance_by_score_PUD_UD}, Appendix \ref{app:performance_of_random_parser_PUD10_SUD} and Appendix \ref{app:performance_of_random_parser_without_sentence_length_limit}).
}
{
(Table \ref{tab:performance_by_distribution_PUD10_UD}, Table \ref{tab:performance_by_score_PUD_UD} and Appendix \ref{app:performance_of_random_parser_without_sentence_length_limit}).
}
Therefore, evaluation without gold standard in other species is {\em a priori} easier than in ours.

At first glance, a reader might conclude that it is not worth parsing non-human primate vocal and gestural sequences due to the high accuracy of a random parser. However, there are several reasons why parsing has value for these sequences.
First, it is worth recognizing that a null model is not necessarily true just because it produces some data. A parser may have the same expected accuracy as the random parser but lower variance. Similarly, failing to reject the null hypothesis with some statistic does not imply that the null hypothesis is correct. We expect that a good-enough parser, namely one that is statistically informed \cite{Marecek2016a,Han2020a}, performs better than the random parser. 
Second, randomness is a general approach to reality that also encompasses  determinism. A deterministic model is particular case of random model such that the probabilities are binary, namely zero or one. A random parser is indeed deterministic when $n = 1$ or $n = 2$, where it always produces the same tree structure. Crucially,  the random parser is more deterministic in non-human primate sequences than in human sentences because of the fast-decaying distribution of sequence lengths of non-human primates. Therefore, the high chance that the random parser yields a correct parse implies that a good-enough parser is channeled to yield a good parse in non-human primates.
 
\subsection{The feasibility of training a parser}

Here we have focused on the feasibility of evaluation. However, there is still another critical question for future research, i.e. the feasibility of training a parser on the sequences that non-human primates produce. A prerequiste is that the sequences are not random in the sense of having some statistical structure. In gelada vocal sequences and chimpanzee gesture sequences, previous units reduce the uncertainty about subsequent units \cite{Gustison2016a,Gustison2017a,Mielke2024a}. \footnote{For geladas in \citet{Gustison2017a}, see Chapter 6, Figure 6.S2. } The same applies to a wide range of other species \cite{Ferrer2005h, Ferrer2012c, Kershenbaum2014a}. Therefore, the primary challenge for training a parser is the low number of sequences available, i.e. of the order several thousands (Table \ref{tab:summary}). 
We speculate that unsupervised learning in non-human primates should be easier than one may initially expect, partly due to the same reasons why evaluation turns out to be easier than expected {\em a priori}. A detailed explanation follows below. 

For human languages, several methods have been developed to constrain the learning space. First, models are often trained with sentences whose tokens are part-of-speech tags (unlexicalized approach) instead of word tokens (lexicalized approach) so that unsupervised learning is easier \cite[Section 5.3]{Han2020a}. 
For instance, the classic Penn Treebank features 36 POS tags \footnote{\url{https://www.ling.upenn.edu/courses/Fall_2003/ling001/penn_treebank_pos.html}} while Universal Dependencies features 17 POS tags. \footnote{\url{https://universaldependencies.org/u/pos/}}
In non-human primates, the number of types is small, 
typically just a couple of dozen in vocal repertoires, e.g., the number of call types produced by male geladas \cite{Gustison2016a} and chimpanzees is 12 in both cases \cite{Girard-Buttoz2022a}, and $\sim 100-150$ in gestural repertoires \cite{Mielke2024b,Grund2025a}.
Therefore, the repertoires of non-human primates have a similar order to the number of part-of-speech tags, which should ease unsupervised learning.
Second, curriculum learning has been used to enhance performance by employing  shorter sentences first \cite{Spitkovsky2010a}. The utility of that kind of approach has been analyzed theoretically \cite{Tu2011a}. 
In non-human primate sequences, the short ones, are the majority (Table \ref{tab:max_sequence_length_in_primates} and Figure \ref{fig:sequence_length_distribution}). Most sequences in the training set will be such that they have only one possible tree, as sequences of length 2 are the most frequent (leaving aside ``sequences'' of length one). 
We conclude that non-human primate sequences have built-in characteristics that ease unsupervised dependency learning and that those characteristics should compensate for the scarcity of data.

\subsection{Conclusion}

The arguments above lead to the following unsupervised parsing methodology for sequences produced by a non-human species
\begin{itemize}
   \item 
   Prerequisites for successful training. Check that there is some statistical structure in the sequences \cite{Ferrer2012c,Mielke2024a}.
   \item 
   Prerequisites for successful training and evaluation. Check that the sequence length distribution decays fast (exponentially). 
   \item 
   Training. Train an unsupervised parser on the sequences \cite{Marecek2012a,Han2020a}. 
   \item 
   Parser validation. Check that the parser is a good-enough parser by checking that it is able to retrieve dependency structures of a quality greater than expected by chance according to some objective function. For instance, by checking that the mutual information of the dependencies is larger than expected for the same parser when trained on a random shuffling of the sequences \cite{Yuret1998}. 
   \item 
   Parser evaluation. Obtain lower bounds of the performance of the good-enough parser by means of $\beta$, the expected value of proportion of correct edges for a random parser. 
   $\beta$ can be $\E[Q]$ or $\E[P_c^e]$.  
   Let $\beta_l$ be the proportion of correct edges retrieved by the random parser on sequences of length $l$ or greater. We know that $\beta_2$ can be a large number while $\beta_3$ will be a smaller number. If $\beta_2$ is large (say $>84\%$ as for chimpanzee gestures according to $\E[Q]$), we can conclude that the performance of the good-enough parser will be high and that its performance on sequences of 3 or greater is likely to be higher than $\beta_3$ 
(say higher than $56\%$ as for chimpanzee gestures)     
   because of the large number of sequences of length 2 reflected in $\beta_2$, which implies very accurate dependency information for training. 
\end{itemize}

\appendix
\gdef\thesection{\Alph{section}} 
\gdef\thesubsection{\thesection.\arabic{subsection}}

\renewcommand{\theHsection}{\thesection.\arabic{section}}
\renewcommand{\theHfigure}{\thesection.\arabic{figure}}
\renewcommand{\theHtable}{\thesection.\arabic{table}}
\renewcommand{\theHequation}{\thesection.\arabic{equation}}

\clearpage

\appendixsection{Probability distributions, expectations and likelihood}
\label{app:probability_distributions_and_expectations}

Here we consider the expectation of $\phi(n)$, a function of $n$, which is,  
\begin{equation*}
\E[\phi(n)] = \sum_{n=n_{min}}^\infty p(n | n \geq n_{min}) \phi(n),
\end{equation*}
assuming that $n$ follows a certain distribution. 

\begin{property}
\label{prop:expectations_for_uniform_distribution}
If the distribution of $n$ is uniform with parameters $n_{min}$ and $n_{max}$ then 
\begin{equation*}
\E[n] = \frac{n_{min} + n_{max}}{2}
\end{equation*}
and
\begin{equation}
\label{eq:expectation_of_reciprocal_uniform_distribution}
\E\left[\frac{1}{n}\right] = \frac{1}{n_{max} - n_{min} + 1} \left[G(n_{max}) - G(n_{min} - 1)\right].
\end{equation}
\end{property}

\begin{proof}
$\E[n]$ is well-known. 
\begin{align*}
\E\left[\frac{1}{n}\right] & = \frac{1}{n_{max} - n_{min} + 1} \sum_{n=n_{min}}^{n_{max}} \frac{1}{n} \commentinequation{Eq. \ref{eq:uniform_distribution}}\\ 
     & = \frac{1}{n_{max} - n_{min} + 1} \left[G(n_{max}) - G(n_{min} - 1)\right],
\end{align*}
where $G(n_{max})$ is the harmonic number of $n_{max}$, defined as
\begin{equation*}
G(n_{max}) = \sum_{n = 1}^{n_{max}} \frac{1}{n}.
\end{equation*}
\end{proof}
It is well-known that
\begin{equation*}
G(n_{max}) \approx \log n_{max} + \gamma,
\end{equation*}
where $\gamma = 0.5772...$ is the Euler-Mascheroni constant. 
Hence Eq. \ref{eq:expectation_of_reciprocal_uniform_distribution} can be approximated as 
\begin{equation*}
\E\left[\frac{1}{n}\right] \approx \frac{1}{n_{max} - n_{min} + 1} \left[\log(n_{max}) + \gamma - G(n_{min} - 1)\right].
\end{equation*}
for small $n_{min}$ and sufficiently large $n_{max}$.

Recall a useful well-known property of the geometric series.
\begin{property}
For $r \in (0, 1)$, 
\label{prop:useful_geometric_series}
\begin{equation*}
A = \sum_{n=n_{min}}^{\infty} r^n = \frac{r^{n_{min}}}{1 - r}.
\end{equation*}
\end{property}
\begin{proof}
The result follows after some algebra from noting that 
\begin{equation*}
r A = A - r^{n_{min}}.
\end{equation*}
\end{proof}
The next property will help us to calculate $\E[1/n]$ for the geometric distribution.

\begin{property}
\label{prop:sum_of_infinite_series}
Consider $0 \leq r < 1$. Then
\begin{equation*}
\sum_{n=n_{min}}^\infty \frac{r^n}{n} = - \log(1 - r) - \sum_{n=1}^{n_{min} - 1} \frac{r^n}{n}.
\end{equation*}
\end{property}
\begin{proof}
The summation 
\begin{equation*}
\sum_{n=n_{min}}^\infty \frac{r^n}{n} \\
\end{equation*}
with $0 \leq r < 1$ converges because 
\begin{equation*}
\sum_{n=n_{min}}^\infty r^n \\
\end{equation*}
converges (recall Property \ref{prop:useful_geometric_series}).
Notice that
\begin{equation*}
\sum_{n=n_{min}}^\infty \frac{r^n}{n} = \sum_{n=1}^\infty \frac{r^n}{n} - \sum_{n=1}^{n_{min} - 1} \frac{r^n}{n}.
\end{equation*}
In turn,
\begin{align*}
\sum_{n=1}^\infty \frac{r^n}{n} & = \sum_{n=1}^\infty  \int_0^r t^{n-1} dt \\
     & = \int_0^r \left(\sum_{n=1}^\infty  t^{n-1}\right) dt \commentinequation{swapping integration and summation}\\
     & = \int_0^r \frac{1}{1 - t} dt \commentinequation{$0 < t < 1$} \commentinequation{Property \ref{prop:useful_geometric_series} with $n_{min} = 0$}\\
     & = - \log (1 - r) 
\end{align*}
Hence Eq. \ref{eq:expectation_of_reciprocal_geometric_distribution}.
The swapping of integration and summation above is justified by the fact that the series $\sum_{n=1}^\infty t^{n-1}$ converges uniformly on the interval $[0,t]$ if $t<1$. 

\end{proof}

\begin{property}
\label{prop:expectations_for_geometric_distribution}
If $n$ follows a geometric distribution with parameters $q$ and $n_{min}$, then 
\begin{equation}
\E[n] = n_{min} - 1 + \frac{1}{q} 
\label{eq:expectation_displaced_geometric_distribution}
\end{equation}
and
\begin{equation}
\label{eq:expectation_of_reciprocal_geometric_distribution}
\E\left[\frac{1}{n}\right] = - q \left(\frac{\log q}{(1 - q)^{n_{min}}} + \sum_{n=1}^{n_{min} - 1} \frac{(1 - q)^{n - n_{min}}}{n} \right).
\end{equation}
\end{property}

\begin{proof}
$n$ follows a geometric distribution defined by $p(n | n \geq n_{min})$ with parameters $q$ and $n_{min}$ (Eq. \ref{eq:geometric_distribution}) and hence $n$ takes integer values on $[n_{min}, \infty)$. 
Eq. \ref{eq:expectation_displaced_geometric_distribution} is a well-known result \cite{Park2023a}.
Besides,
\begin{align*}
\E\left[\frac{1}{n}\right] & = \frac{q}{(1 - q)^{n_{min}}} \sum_{n=n_{min}}^\infty \frac{(1 - q)^n}{n}. \\
\end{align*}
Property \ref{prop:sum_of_infinite_series} with $r = 1 - q$ gives 
\begin{equation*}
\E\left[\frac{1}{n}\right] = - \frac{q}{(1 - q)^{n_{min}}}\left(\log q + \sum_{n=1}^{n_{min} - 1} \frac{(1 - q)^n}{n} \right).
\end{equation*}
Hence Eq. \ref{eq:expectation_of_reciprocal_geometric_distribution}.
\end{proof}

Finally, we wish to estimate the parameters of a geometric distribution of sequence lengths by maximizing the likelihood. Suppose we have $S$ sequences of length $n_{min}$ or greater. Then the likelihood of the observed sequence lengths is 
\begin{equation}
L = \prod_{i=1}^S p(n_i).
\label{eq:likelihood}
\end{equation}

The following property provides well-known biased parameter estimators.
\begin{property}
\label{prop:maximum_likelihood_parameters_geometric_distribution}
Suppose a geometric distribution with parameters $q$ and $n_{min}$.
Suppose a sample formed by $S$ sequences of length $n_{min}$ or greater and $T$ tokens in total.
The maximum likelihood estimator for $n_{min}$ is the minimum sequence length in the sample.
The maximum likelihood estimator for $q$ is
\begin{align}
\hat{q} & = \frac{S}{T - S(n_{min} - 1)} \label{eq:maximum_likeligood_estimator_of_q} \\ 
        & = (\left<n\right> + 1 - n_{min})^{-1}. \label{eq:maximum_likeligood_estimator_of_q_mean_sequence_length}
\end{align}
Hence
\begin{equation*}
\hat{q} = 
    \left\{
        \begin{array}{ll}
        1/(\left<n\right> - 1) & \mbox{~for $n_{min} = 0$} \\
        1/\left<n\right>       & \mbox{~for $n_{min} = 1$} \\
        1/(\left<n\right> + 1) & \mbox{~for $n_{min} = 2$}.
        \end{array}
    \right.      
\end{equation*}
\end{property}
\begin{proof}
Eq. \ref{eq:maximum_likeligood_estimator_of_q_mean_sequence_length} is well-known \cite{Park2023a}. From it all other equations follow easily but we wish to be clear on how $n_{min}$ has to visit and that implies revising the whole derivation process. 

Then log-likelihood is (Eq. \ref{eq:likelihood})
\begin{equation}
{\cal L} = \log L = \sum_{i=1}^S\log p(n_i).
\label{eq:log_likelihood}
\end{equation}
Applying Eq. \ref{eq:geometric_distribution} to Eq. \ref{eq:log_likelihood}, we get 
\begin{align*}
{\cal L} & = \sum_{i=1}^S \log q + \sum_{i=1}^S (n_i - n_{min}) \log (1-q) \\
         & = S \log q + \log (1-q) \left(\sum_{i=1}^S (n_i - n_{min})\right) \\
         & = S \log q + (T - S n_{min}) \log (1-q).
\end{align*}
It is easy to see that ${\cal L}$ is a decreasing function of $n_{min}$.
Therefore, ${\cal L}$ is maximized when $n_{min}$ is set to the minimum sequence length in the sample. 

Setting $\partial {\cal L} / \partial q = 0$,
we obtain Eq. \ref{eq:maximum_likeligood_estimator_of_q} after some algebra. 
Noting that $\left<n \right> = T/S$ we obtain Eq.  \ref{eq:maximum_likeligood_estimator_of_q_mean_sequence_length}.
\end{proof}
$\hat{q}$ is a biased estimator of $q$ \cite{Park2023a}.  
The following estimator fixes the problem. 

\begin{property}[\citet{Park2023a}]
\label{prop:minimum_variance_unbiased_parameter_geometric_distribution}
The minimum variance unbiased (MVU) estimator for $q$ is 
\begin{align}
\hat{q} & = \frac{S - 1}{T - S(n_{min} - 1) - 1} \label{eq:MVU_estimator_of_q} \\ 
        & = \frac{(S - 1)/S}{\left<n \right> + 1 - n_{min} - 1/S}. \label{eq:MVU_estimator_of_q_mean_sequence_length}
\end{align}
Hence
\begin{eqnarray*}
\hat{q} 
    & = 
    \left\{
        \begin{array}{ll}
        \frac{(S - 1)/S}{\left<n\right> - 1 - 1/S)} & \mbox{~for $n_{min} = 0$} \\
        \frac{(S - 1)/S}{\left<n\right> - 1/S}      & \mbox{~for $n_{min} = 1$} \\
        \frac{(S - 1)/S}{\left<n\right> + 1 - 1/S}  & \mbox{~for $n_{min} = 2$}.
        \end{array}
    \right.      
\end{eqnarray*}
\end{property}

\appendixsection{The expected proportion of correct trees}
\label{app:expected_proportion_of_correct_trees}

A sequence of length $n$ has $n^{n - 2}$ labeled trees \cite{Cayley1889a}. The probability of obtaining the correct tree by picking one of them uniformly at random is 
\begin{equation}
p(c_i = 1 | n) = 1/n^{n-2} = n^{2-n}.
\label{eq:random_tree_probability}
\end{equation}
The probability of choosing the correct tree at random in a sequence of three units is $1/3$; in a sequence of four units, this probability is $1/16$.

The expected value of $P_c^t$ given $S(n_{min})$ is
\begin{align}
\E[P_c^t] & = \frac{1}{S(n_{min})}\E[S_c(n_{min})] \commentinequation{Eq. \ref{eq:proportion_of_correct_trees}} \nonumber \\
          & = \frac{1}{S(n_{min})}\sum_{\substack{i=1 \\ n_i \geq n_{min}}}^S \E\left[c_i\right] \commentinequation{Eq. \ref{eq:number_of_correct_trees} and linearity of expectation}  \nonumber \\
          & =  \E[c_i]. \commentinequation{identically distributed $c_i$'s} 
          \label{eq:raw_expected_proportion_of_correct_trees}
\end{align}

As $c_i$ is an indicator variable, $\E[c_i]$ becomes the probability that a uniformly random labeled tree matches the correct tree of an arbitrary sequence of length $n_{min}$ or greater, namely
\begin{align*}
\E[c_i]          & = \sum_{n=n_{min}}^{\infty} p(n | n \geq n_{min}) p(c_i = 1 | n) \commentinequation{law of total expectation} \\
                 & = \sum_{n=n_{min}}^{\infty} p(n | n \geq n_{min}) n^{2 - n}. \commentinequation{Eq. \ref{eq:random_tree_probability}}
\end{align*}
Then Eq. \ref{eq:raw_expected_proportion_of_correct_trees} becomes
\begin{equation}
\E[P_c^t] = \sum_{n=n_{min}}^{\infty} p(n | n \geq n_{min}) n^{2 - n}. \label{eq:expected_proportion_of_correct_trees}
\end{equation}
For the uniform distribution above (Eq. \ref{eq:uniform_distribution}),  
\begin{equation*}
\E[P_c^t] = \frac{1}{n_{max} - n_{min} + 1} \sum_{n=n_{min}}^{n_{max}} n^{2 - n}.
\end{equation*}
For the 2-parameter geometric distribution above (Eq. \ref{eq:geometric_distribution}) 
\begin{equation*}
\E[P_c^t] = q (1-q)^{- n_{min}} \sum_{n=n_{min}}^{n_{max}} (1-q)^n n^{2 - n}.
\end{equation*}
For the empirical distribution
, Eq. \ref{eq:empirical_sequence_length_probability}
and Eq. \ref{eq:expected_proportion_of_correct_trees} give
\begin{equation*}
\E[P_c^t] = \frac{1}{S(n_{min})} \sum_{n = n_{min}}^{n_{max}}f(n) n^{2 - n}.
\end{equation*}
In the last case, $\E[P_c^t]$ is the expected proportion of correct random trees when the length of each sequence is provided. 

\appendixsection{The expected proportion of correct edges}
\label{app:expected_proportion_of_correct_edges}

First, we assume that ${\cal F}$ consists of a single tree of $n$ vertices. Then $m = n - 1$ and $m_c$ becomes the number of 
edges of a tree that are correct and then
\begin{equation*}
P_c^e = \frac{m_c}{n - 1}    
\end{equation*}
and 
\begin{equation*}
\E[P_c^e| n] = \frac{1}{n - 1}\E[m_c | n],
\end{equation*}
where $\E[m_c | n]$ is the expected size of the intersection between the correct tree and the tree produced by the parser given their size $n$. It is easy to see that, $\E[m_c | n] = 1$ when $n = 2$.
For $n > 2$, recall that we have assumed that all trees of same size have equal chance of being the correct tree.
Then, it is easy to see that $\E[m_c] = \frac{4}{3}$ when $n=3$. The point is that there are three labeled trees. When selecting a pair of those trees at random: the trees can be the same with probability 1/3, which gives $m_c = 2$ or the trees are different with probability 2/3, which gives $m_c = 1$. Then 
\begin{equation*}
\E[m_c] = \frac{1}{3}2 + \frac{2}{3}1 = \frac{4}{3}.     
\end{equation*}
$\E[m_c]$ is known as the value of the tree intersection game, namely the number of edges shared by two random trees \cite{London2023a}. If trees are uniformly random labeled trees, one has \cite[Observation 3]{London2023a}
\begin{equation}
\E[m_c | n ] = \frac{2(n - 1)}{n}
\label{eq:expected_number_of_intersecting_edges_over_single_tree}
\end{equation}
for $n \geq  1$.
Hence
\begin{equation}
\E[P_c^e | n] = \frac{2}{n} \label{eq:expected_proportion_of_intersecting_edges_over_single_tree}
\end{equation}
for $n > 1$. When $n = 1$, $\E[P_c^e]$ is undefined.   
The expected proportion of correct edges retrieved when $n=2$ is obviously 1, when $n=3$ it is $2/3$ and when $n=4$ it is $1/2$.
In the original setting of \citet{London2023a}, one selects two random spanning trees from a graph $G$. In our application of their framework, $G$ is a complete graph of $n$ vertices and one of the spanning trees has the role of the spanning tree and the other spanning tree has the role of the tree retrieved by the random parser. 

Now we return to the general case, where ${\cal F}$ may be formed by one or more trees. The two ways of measuring the performance of the parser above parallel the two ways of measuring the clustering coefficient in network science \cite[Chapter 7]{Newman2010a}: one is the proportion of paths of lengths two that are closed (the counterpart of $P_c^e$) and the other is the average local proportion over the paths formed by the neighbors of a vertex (the counterpart of $Q$). 

We aim to calculate the expectation of $P_c^e$ and that of $Q$. We have 
\begin{align}
\E[m_{c,i}] & = \sum_{n=n_{min}}^\infty p(n | n \geq n_{min}) \E[m_c | n ] \commentinequation{law of total expectation} \nonumber \\
             & = 2 \sum_{n=n_{min}}^\infty p(n | n \geq n_{min})\left(1 - \frac{1}{n}\right) \commentinequation{Eq. \ref{eq:expected_number_of_intersecting_edges_over_single_tree}}
             \label{eq:expected_number_of_intersecting_edges_over_single_tree_unknown_tree_size}
\end{align}
and
\begin{align}
\E[P_{c,i}^e] & = \sum_{n=n_{min}}^\infty p(n | n \geq n_{min}) \E[m_c | n ] \commentinequation{law of total expectation} \nonumber \\
             & = 2 \sum_{n=n_{min}}^\infty \frac{p(n | n \geq n_{min})}{n}. \commentinequation{Eq. \ref{eq:expected_proportion_of_intersecting_edges_over_single_tree}}
             \label{eq:expected_proportion_of_intersecting_edges_over_single_tree_unknown_tree_size}
\end{align}
Then 
\begin{align}
\E[Q]  & = \frac{1}{S(n_{min})} \sum_{\substack{i=1 \\ n_i \geq n_{min}}}^{S} \E[P_{c,i}^e] \commentinequation{linearity of expectation} \nonumber \\
       & = \E[P_{c,i}^e] \nonumber \\
       & = 2 \sum_{n=n_{min}}^\infty \frac{p(n | n \geq n_{min})}{n}. \commentinequation{Eq. \ref{eq:expected_proportion_of_intersecting_edges_over_single_tree_unknown_tree_size}} \label{eq:expected_proportion_of_intersecting_edges_over_trees}            
\end{align}

For the empirical distribution, Eq. \ref{eq:empirical_sequence_length_probability} and \ref{eq:expected_proportion_of_intersecting_edges_over_trees} yield
\begin{equation*}
\E[Q] = \frac{2}{H(n_{min})},
\end{equation*}
where $H(n_{min})$ is the harmonic mean of sentence lengths that are $n_{min}$ or larger, that is
\begin{align}
H(n_{min}) & = \frac{1}{S(n_{min})} \sum_{n = n_{min}}^{n_{max}} \frac{f(n)}{n} \nonumber \\
& = \frac{1}{S(n_{min})} \sum_{\substack{i=1 \\ n_i \geq n_{min}}}^S \frac{1}{n_i}. \label{eq:harmonic_mean}
\end{align}

Besides, we have 
\begin{align}
\E[P_c^e] & = \frac{1}{m(n_{min})}\E\left[m_c(n_{min})\right] \commentinequation{Eq. \ref{eq:proportion_of_correct_edges} and $m(n_{min})$ is constant} \nonumber \\
            & = \frac{1}{m(n_{min})} \sum_{\substack{i=1 \\ n_i \geq n_{min}}}^S \E[m_{c,i}] \commentinequation{Eq. \ref{eq:total_number_of_edges} and linearity of expectation} \nonumber \\
            & = \frac{2 S(n_{min})}{m(n_{min})} \sum_{n=n_{min}}^\infty p(n | n \geq n_{min})\left(1 - \frac{1}{n}\right) \commentinequation{Eq. \ref{eq:expected_number_of_intersecting_edges_over_single_tree_unknown_tree_size}} \label{eq:expected_proportion_of_intersecting_edges_over_edges}             
\end{align}
            
For the empirical distribution, Eq. \ref{eq:empirical_sequence_length_probability} and Eq. \ref{eq:expected_proportion_of_intersecting_edges_over_edges} yield 
\begin{align*}
\E[P_c^e] & = \frac{2}{m(n_{min})} \sum_{n = n_{min}}^{n_{max}} f(n) \left(1 - \frac{1}{n}\right) \commentinequation{Eq. \ref{eq:expected_number_of_intersecting_edges_over_single_tree}} \\
             & = \frac{2}{m(n_{min})} [S(n_{min}) - H(n_{min})]. \commentinequation{Eq. \ref{eq:harmonic_mean}}
\end{align*}

\appendixsection{The theoretical performance of the random parser}
\label{app:theoretical_performance_of_random_parser}

We examine the performance of the random parser as a direct function of the parameters of the distribution. In increase of $n_{max}$ (in the uniform distribution) or a reduction of $q$ (in the geometric distribution) leads to longer sequences.   

\begin{figure}
    \centering
    \includegraphics[width=0.9\linewidth]{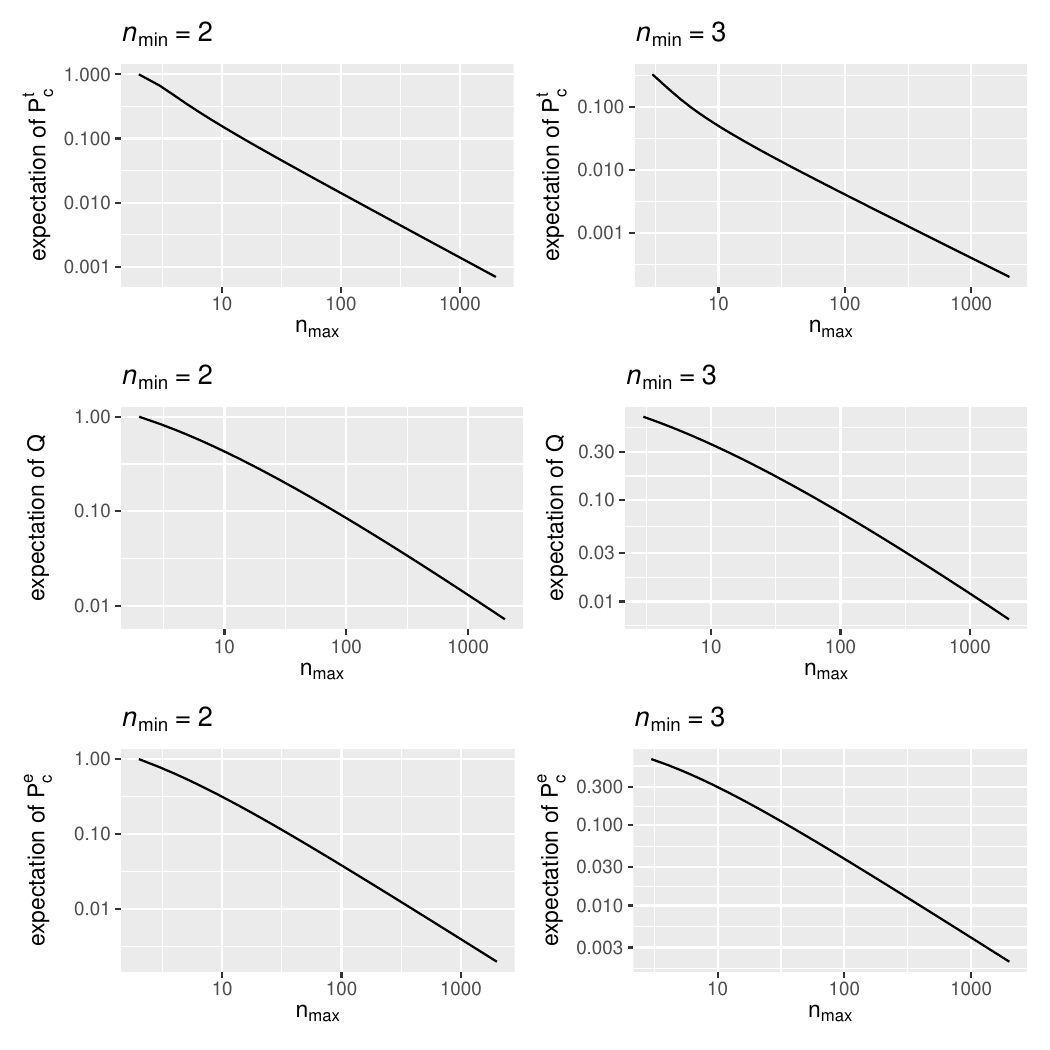}
    \caption{
    The performance of a random parser as a function of the parameter $n_{max}$ of a uniform  distribution in the interval $[1, n_{max}]$ (Eq. \ref{eq:uniform_distribution}).     
    Three performance scores are considered: $\E[P_c^t]$, the expected proportion of correct trees (top), $\E[Q]$, the average expected proportion of correct edges per sequence (middle) and $\E[P_c^e]$, the expected overall proportion of correct edges (bottom). On top each subfigure, $n_{min}$ indicates the minimum sequence length considered to measure the performance of the parser.
    }
    \label{fig:performance_uniform_geometric_distribution}
\end{figure}

\begin{figure}
    \centering
    \includegraphics[width=0.9\linewidth]{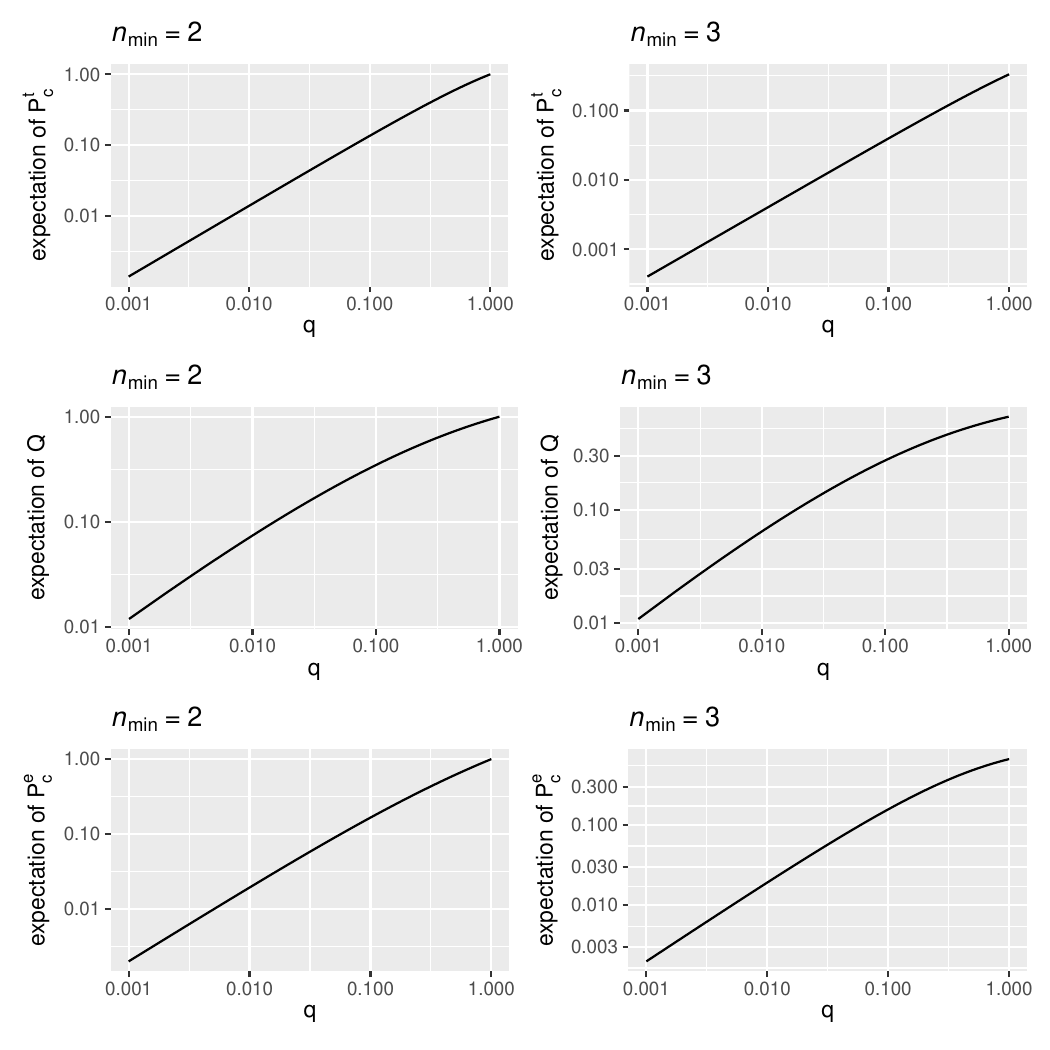}
    \caption{The performance of a random parser as a function of the parameter $q$ of a geometric distribution on $[1, \infty)$ (Eq. \ref{eq:geometric_distribution}). The format is the same as in Fig. \ref{fig:performance_uniform_geometric_distribution}.    
    }
    \label{fig:performance_geometric_distribution}
\end{figure}

As expected from the Section \ref{subsec:theoretical_performance_of_random_parser}, performance decreases as the parameters $n_{max}$ of the uniform distribution increases 
(Fig. \ref{fig:performance_uniform_geometric_distribution})
while it increases as the parameter $q$ of the geometric distribution increases (Fig. \ref{fig:performance_geometric_distribution}).

\appendixsection{Computation of the expected performance of the random parser for the geometric distribution}
\label{app:calculation_of_the_expected_performance_of_random_parser}

We aim to calculate $K$ (Eq. \ref{eq:generic_K_score}) with high numerical precision for the geometric distribution. Consider $\widetilde{K}$ as an approximation of $K$ up to $n_*$, that is
\begin{equation*}
\widetilde{K} = c \sum_{n = n_{min}}^{n_*} p(n | n \geq n_{min}) \phi(n). \commentinequation{Eq. \ref{eq:generic_expectation} and Eq. \ref{eq:generic_K_score}}
\end{equation*}
We define the approximation error of $\widetilde{K}$ as $\epsilon = K - \widetilde{K}$. Notice that $\epsilon \geq 0$ since $\widetilde{K} \leq K$. We have 
\begin{align*}
\epsilon & = c \sum_{n = n_* + 1}^{\infty} p(n | n \geq n_{min}) \phi(n) \commentinequation{Eq. \ref{eq:generic_expectation} and Eq. \ref{eq:generic_K_score}}\\
         & = cq(1-q)^{-n_{min}} \sum_{n = n_*+1}^{\infty} (1-q)^n \phi(n). \\
\end{align*}

We define $\phi_{max}$ as an upper bound of $\phi(n)$ in the interval $[n_* + 1, \infty)$. We also define
$\epsilon_{max}$, an upper bound of $\epsilon$, that is obtained by setting $\phi(n)$ to $\phi_{max}$ and noting that (Property \ref{prop:useful_geometric_series})
\begin{equation*}
\sum_{n = n_* +1}^\infty (1-q)^n = \frac{(1-q)^{n_* + 1}}{q},
\end{equation*}
which yields 
\begin{equation*}
    \epsilon_{max} = c\phi_{max}(1-q)^{n_* + 1 - n_{min}}. \\
\end{equation*}
Rearranging terms one gets
\begin{equation*}
\frac{\epsilon (1- q)^{n_{min} - 1}}{c \phi_{max}} = (1-q)^{n_*}.
\end{equation*}
Taking logarithms on both sides, we get
\begin{equation*}
n^* = \frac{\log\frac{\epsilon}{c \phi_{max}}}{\log (1 - q)} + n_{min} - 1.
\end{equation*}
Changing the base of the logarithm to $1-q$ and truncating (ceiling) to ensure that $\epsilon \leq \epsilon_{max}$, one eventually gets
\begin{equation}
n_* = \left\lceil \log_{1-q} \frac{\epsilon}{c\phi_{max}} + n_{min} - 1 \right\rceil.
\label{eq:n_star}
\end{equation}
The values of $c$ and $\phi_{max}$ depend on the score that one wishes to compute from $K$. However $c \leq 2$ and $\phi(n) \leq 1$ independently of the score. Therefore, a simple upper bound of $n_*$, $n_*'$ is obtained by setting $c = 2$ and $\phi_{max} = 1$ in Eq. \ref{eq:n_star}, which yields
\begin{equation*}
n_*' = \left\lceil \log_{1 - q} \frac{\epsilon}{2} + n_{min} - 1 \right\rceil.
\end{equation*}

\appendixsection{The distribution of sentence length in PUD}
\label{app:sentence_length_distribution}

\begin{figure}
    \centering
    \includegraphics[height=0.9\textheight]{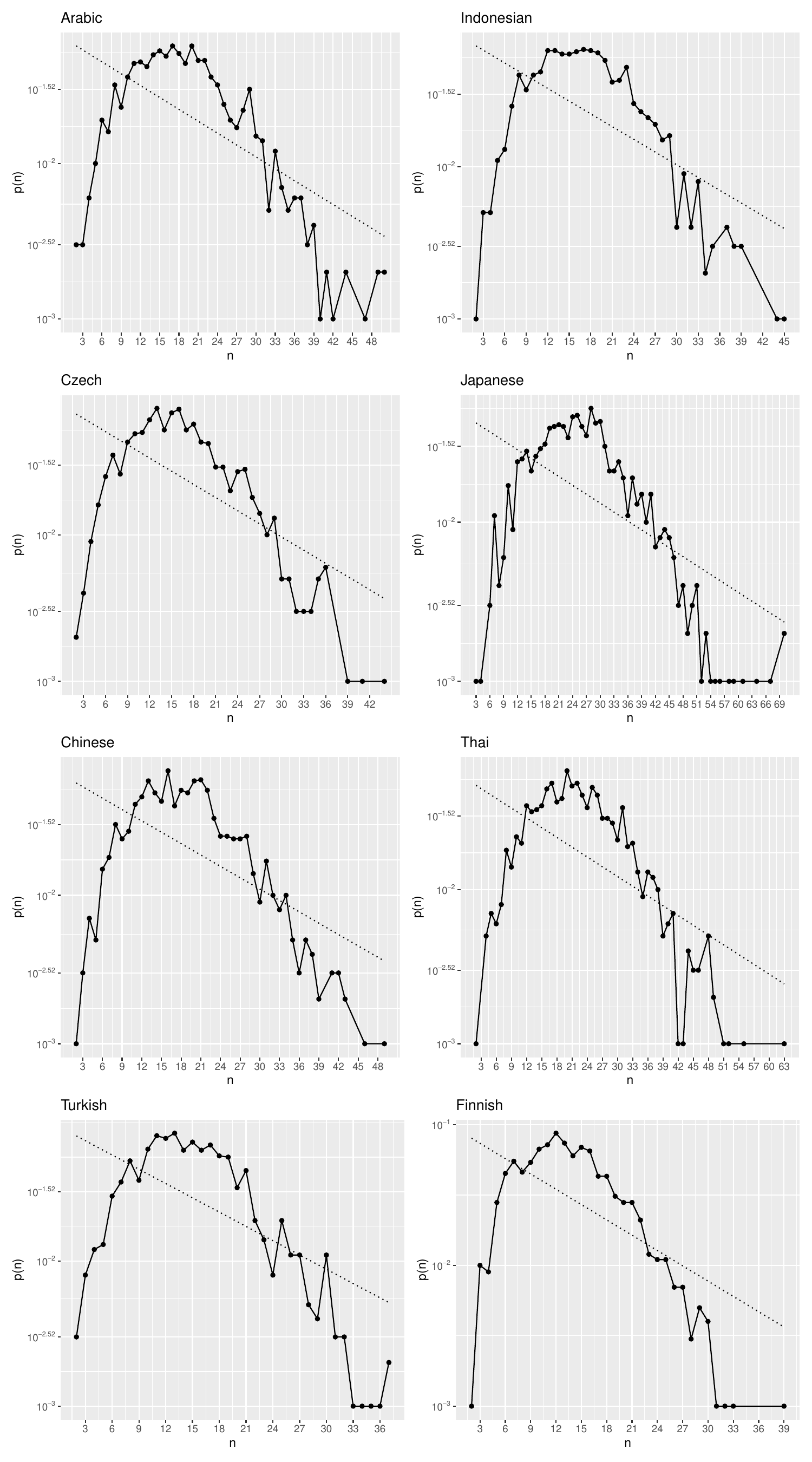}
    \caption{\label{fig:sequence_length_distribution_PUD} The empirical distribution of sentence lengths for selected human languages in the PUD \iftoggle{PSUD}{collection with UD annotation style.}{collection.} On top of each subfigure the languages is shown. Each language corresponds to a distinct linguistic family (Table \ref{tab:PUD_languages}).  
    The dotted line shows the best fit of a 2-parameter geometric distribution (Eq. \ref{eq:geometric_distribution}) with parameters $n_{min}=2$ and 
    $q =\frac{(S - 1)/S}{\left<n\right> + 1 - 1/S}$ where $S$ is the number of sequences and $\left<n\right>$ is the average sequence length
    (Appendix \ref{app:probability_distributions_and_expectations}).
    }
\end{figure}

\iftoggle{PSUD}{
Here we show the distribution of sentence length in PUD with UD annotation style
(Figure \ref{fig:sequence_length_distribution_PUD} and with SUD annotation style (Figure \ref{fig:sequence_length_distribution_PSUD}).
}
{
The distribution of sentence length in PUD is shown in Figure \ref{fig:sequence_length_distribution_PUD}. 
}

\iftoggle{PSUD}{
\begin{figure}
    \centering
    \includegraphics[height=0.95\textheight]{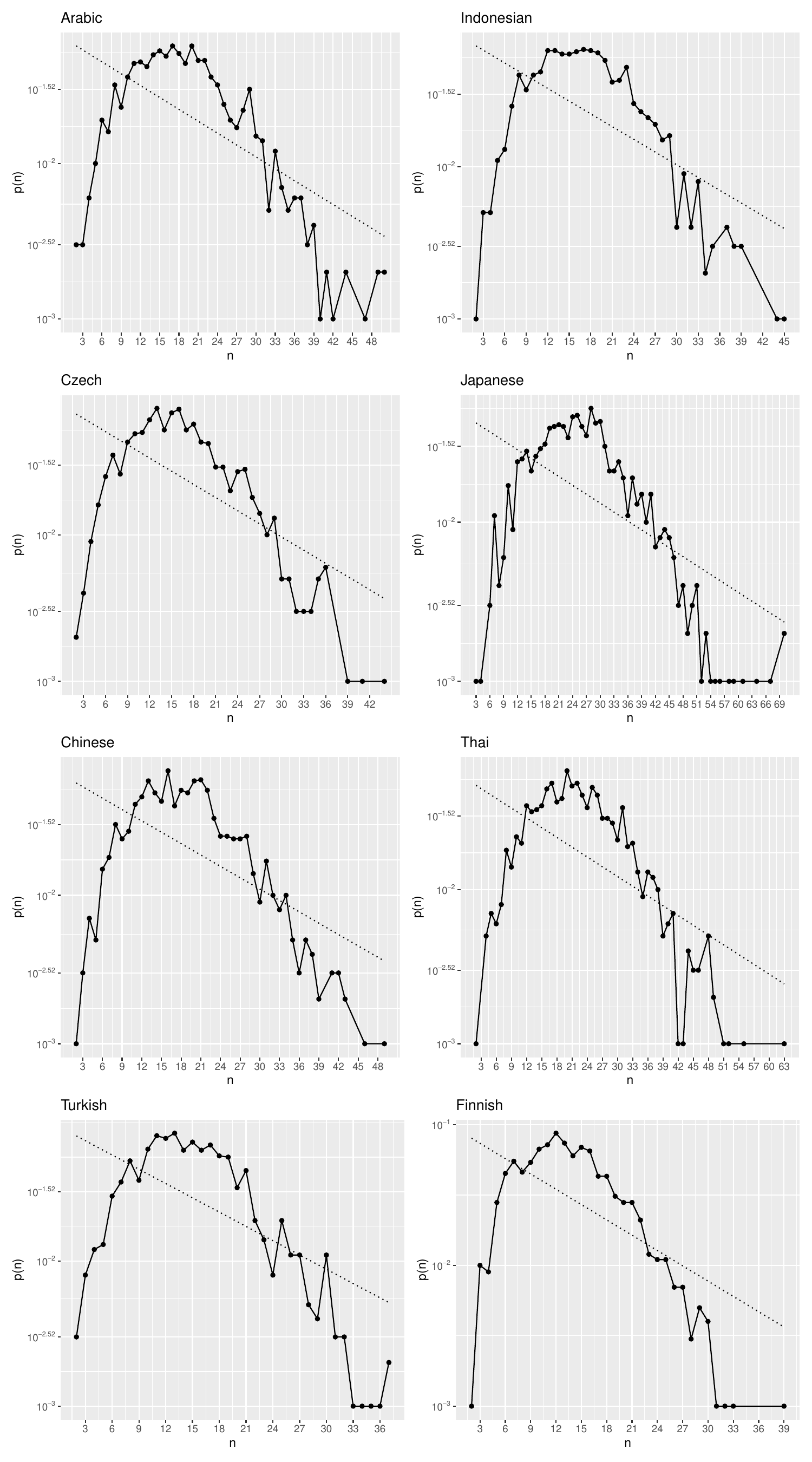}
    \caption{\label{fig:sequence_length_distribution_PSUD} The empirical distribution of sentence lengths for selected human languages in the PUD collection with SUD annotation style. The format is the same as in Fig. \ref{fig:sequence_length_distribution_PUD}. }
\end{figure}
}

\iftoggle{PSUD}{
\appendixsection{Performance in PUD10 with SUD annotation style}
\label{app:performance_of_random_parser_PUD10_SUD}

\begin{table}
    \centering
    \caption{\label{tab:performance_by_score_PUD10_SUD} The performance scores $\E[P_c^t]$, $\E[Q]$, $\E[P_c^e]$ on human languages in the PUD10 collection (sentences of up to length 10 in PUD). The format is the same as in Table \ref{tab:performance_by_score_PUD10_UD}. 
    }    
    {\footnotesize
    \begin{tabular}{rlrrrrrrrrr}
     & & \multicolumn{3}{c}{$\E[P_c^t]$} & \multicolumn{3}{c}{$\E[Q]$} & \multicolumn{3}{c}{$\E[P_c^e]$} \\\cmidrule(lr){3-5} \cmidrule(lr){6-8} \cmidrule(lr){9-11}
    $n_{min}$ & language & u & g & e & u & g & e & u & g & e \\
    \hline
    2 & Arabic & 0.156 & 0.198 & 0.03 & 0.429 & 0.433 & 0.292 & 0.314 & 0.233 & 0.255\\
2 & Chinese & 0.156 & 0.197 & 0.019 & 0.429 & 0.431 & 0.284 & 0.314 & 0.231 & 0.255\\
2 & Czech & 0.156 & 0.198 & 0.019 & 0.429 & 0.432 & 0.286 & 0.314 & 0.233 & 0.255\\
2 & English & 0.156 & 0.227 & 0.009 & 0.429 & 0.467 & 0.275 & 0.314 & 0.263 & 0.254\\
2 & Finnish & 0.156 & 0.202 & 0.016 & 0.429 & 0.437 & 0.289 & 0.314 & 0.237 & 0.26\\
2 & French & 0.156 & 0.223 & 0.012 & 0.429 & 0.463 & 0.276 & 0.314 & 0.259 & 0.251\\
2 & Galician & 0.156 & 0.203 & 0.036 & 0.429 & 0.439 & 0.303 & 0.314 & 0.238 & 0.261\\
2 & German & 0.156 & 0.223 & 0.007 & 0.429 & 0.463 & 0.273 & 0.314 & 0.259 & 0.25\\
2 & Hindi & 0.156 & 0.223 & 0.01 & 0.429 & 0.462 & 0.272 & 0.314 & 0.258 & 0.25\\
2 & Icelandic & 0.156 & 0.223 & 0.01 & 0.429 & 0.463 & 0.274 & 0.314 & 0.259 & 0.25\\
2 & Indonesian & 0.156 & 0.194 & 0.018 & 0.429 & 0.427 & 0.278 & 0.314 & 0.228 & 0.251\\
2 & Italian & 0.156 & 0.198 & 0.018 & 0.429 & 0.432 & 0.286 & 0.314 & 0.232 & 0.256\\
2 & Japanese & 0.156 & 0.206 & 0.009 & 0.429 & 0.443 & 0.257 & 0.314 & 0.241 & 0.238\\
2 & Korean & 0.156 & 0.194 & 0.014 & 0.429 & 0.427 & 0.277 & 0.314 & 0.228 & 0.251\\
2 & Polish & 0.156 & 0.193 & 0.023 & 0.429 & 0.426 & 0.284 & 0.314 & 0.227 & 0.249\\
2 & Portuguese & 0.156 & 0.193 & 0.023 & 0.429 & 0.426 & 0.284 & 0.314 & 0.227 & 0.249\\
2 & Russian & 0.156 & 0.195 & 0.023 & 0.429 & 0.429 & 0.283 & 0.314 & 0.23 & 0.252\\
2 & Spanish & 0.156 & 0.2 & 0.032 & 0.429 & 0.434 & 0.295 & 0.314 & 0.234 & 0.257\\
2 & Swedish & 0.156 & 0.195 & 0.013 & 0.429 & 0.428 & 0.277 & 0.314 & 0.229 & 0.252\\
2 & Thai & 0.156 & 0.192 & 0.017 & 0.429 & 0.425 & 0.279 & 0.314 & 0.226 & 0.25\\
2 & Turkish & 0.156 & 0.2 & 0.027 & 0.429 & 0.434 & 0.292 & 0.314 & 0.234 & 0.256\\
3 & Arabic & 0.051 & 0.058 & 0.01 & 0.357 & 0.334 & 0.277 & 0.299 & 0.216 & 0.253\\
3 & Chinese & 0.051 & 0.058 & 0.012 & 0.357 & 0.332 & 0.278 & 0.299 & 0.214 & 0.254\\
3 & Czech & 0.051 & 0.058 & 0.01 & 0.357 & 0.333 & 0.279 & 0.299 & 0.215 & 0.254\\
3 & English & 0.051 & 0.067 & 0.009 & 0.357 & 0.357 & 0.275 & 0.299 & 0.24 & 0.254\\
3 & Finnish & 0.051 & 0.06 & 0.013 & 0.357 & 0.337 & 0.287 & 0.299 & 0.219 & 0.26\\
3 & French & 0.051 & 0.066 & 0.012 & 0.357 & 0.354 & 0.276 & 0.299 & 0.237 & 0.251\\
3 & Galician & 0.051 & 0.06 & 0.016 & 0.357 & 0.338 & 0.289 & 0.299 & 0.22 & 0.259\\
3 & German & 0.051 & 0.066 & 0.007 & 0.357 & 0.354 & 0.273 & 0.299 & 0.237 & 0.25\\
3 & Hindi & 0.051 & 0.066 & 0.01 & 0.357 & 0.353 & 0.272 & 0.299 & 0.237 & 0.25\\
3 & Icelandic & 0.051 & 0.066 & 0.01 & 0.357 & 0.354 & 0.274 & 0.299 & 0.237 & 0.25\\
3 & Indonesian & 0.051 & 0.057 & 0.012 & 0.357 & 0.33 & 0.274 & 0.299 & 0.211 & 0.25\\
3 & Italian & 0.051 & 0.058 & 0.009 & 0.357 & 0.333 & 0.279 & 0.299 & 0.215 & 0.255\\
3 & Japanese & 0.051 & 0.061 & 0.009 & 0.357 & 0.34 & 0.257 & 0.299 & 0.223 & 0.238\\
3 & Korean & 0.051 & 0.057 & 0.01 & 0.357 & 0.33 & 0.274 & 0.299 & 0.212 & 0.251\\
3 & Polish & 0.051 & 0.057 & 0.014 & 0.357 & 0.329 & 0.277 & 0.299 & 0.211 & 0.248\\
3 & Portuguese & 0.051 & 0.057 & 0.014 & 0.357 & 0.329 & 0.277 & 0.299 & 0.211 & 0.248\\
3 & Russian & 0.051 & 0.057 & 0.013 & 0.357 & 0.331 & 0.276 & 0.299 & 0.213 & 0.251\\
3 & Spanish & 0.051 & 0.059 & 0.011 & 0.357 & 0.335 & 0.28 & 0.299 & 0.217 & 0.255\\
3 & Swedish & 0.051 & 0.057 & 0.007 & 0.357 & 0.331 & 0.273 & 0.299 & 0.212 & 0.252\\
3 & Thai & 0.051 & 0.056 & 0.005 & 0.357 & 0.328 & 0.27 & 0.299 & 0.21 & 0.248\\
3 & Turkish & 0.051 & 0.059 & 0.015 & 0.357 & 0.335 & 0.283 & 0.299 & 0.217 & 0.255\\
 
    \end{tabular}
    }
\end{table}

\begin{table}
    \centering
\caption{\label{tab:performance_by_distribution_PUD10_SUD} The performance scores $\E[P_c^t]$, $\E[Q]$, $\E[P_c^e]$ on human languages in the PUD10 collection (sentences of up to length 10 in PUD) with SUD annotation style. The format is the same as in Table \ref{tab:performance_by_distribution_PUD10_UD}. }    
    {\footnotesize
    \begin{tabular}{rlrrrrrrrrr}
     & & \multicolumn{3}{c}{uniform} & \multicolumn{3}{c}{geometric} & \multicolumn{3}{c}{empirical} \\\cmidrule(lr){3-5} \cmidrule(lr){6-8} \cmidrule(lr){9-11}
    $n_{min}$ & language &  
    $\E[P_c^t]$ & $\E[Q]$ & $\E[P_c^e]$ &   
    $\E[P_c^t]$ & $\E[Q]$ & $\E[P_c^e]$ &
    $\E[P_c^t]$ & $\E[Q]$ & $\E[P_c^e]$ \\
    \hline
    2 & Arabic & 0.156 & 0.429 & 0.314 & 0.198 & 0.433 & 0.233 & 0.03 & 0.292 & 0.255\\
2 & Chinese & 0.156 & 0.429 & 0.314 & 0.197 & 0.431 & 0.231 & 0.019 & 0.284 & 0.255\\
2 & Czech & 0.156 & 0.429 & 0.314 & 0.198 & 0.432 & 0.233 & 0.019 & 0.286 & 0.255\\
2 & English & 0.156 & 0.429 & 0.314 & 0.227 & 0.467 & 0.263 & 0.009 & 0.275 & 0.254\\
2 & Finnish & 0.156 & 0.429 & 0.314 & 0.202 & 0.437 & 0.237 & 0.016 & 0.289 & 0.26\\
2 & French & 0.156 & 0.429 & 0.314 & 0.223 & 0.463 & 0.259 & 0.012 & 0.276 & 0.251\\
2 & Galician & 0.156 & 0.429 & 0.314 & 0.203 & 0.439 & 0.238 & 0.036 & 0.303 & 0.261\\
2 & German & 0.156 & 0.429 & 0.314 & 0.223 & 0.463 & 0.259 & 0.007 & 0.273 & 0.25\\
2 & Hindi & 0.156 & 0.429 & 0.314 & 0.223 & 0.462 & 0.258 & 0.01 & 0.272 & 0.25\\
2 & Icelandic & 0.156 & 0.429 & 0.314 & 0.223 & 0.463 & 0.259 & 0.01 & 0.274 & 0.25\\
2 & Indonesian & 0.156 & 0.429 & 0.314 & 0.194 & 0.427 & 0.228 & 0.018 & 0.278 & 0.251\\
2 & Italian & 0.156 & 0.429 & 0.314 & 0.198 & 0.432 & 0.232 & 0.018 & 0.286 & 0.256\\
2 & Japanese & 0.156 & 0.429 & 0.314 & 0.206 & 0.443 & 0.241 & 0.009 & 0.257 & 0.238\\
2 & Korean & 0.156 & 0.429 & 0.314 & 0.194 & 0.427 & 0.228 & 0.014 & 0.277 & 0.251\\
2 & Polish & 0.156 & 0.429 & 0.314 & 0.193 & 0.426 & 0.227 & 0.023 & 0.284 & 0.249\\
2 & Portuguese & 0.156 & 0.429 & 0.314 & 0.193 & 0.426 & 0.227 & 0.023 & 0.284 & 0.249\\
2 & Russian & 0.156 & 0.429 & 0.314 & 0.195 & 0.429 & 0.23 & 0.023 & 0.283 & 0.252\\
2 & Spanish & 0.156 & 0.429 & 0.314 & 0.2 & 0.434 & 0.234 & 0.032 & 0.295 & 0.257\\
2 & Swedish & 0.156 & 0.429 & 0.314 & 0.195 & 0.428 & 0.229 & 0.013 & 0.277 & 0.252\\
2 & Thai & 0.156 & 0.429 & 0.314 & 0.192 & 0.425 & 0.226 & 0.017 & 0.279 & 0.25\\
2 & Turkish & 0.156 & 0.429 & 0.314 & 0.2 & 0.434 & 0.234 & 0.027 & 0.292 & 0.256\\
3 & Arabic & 0.051 & 0.357 & 0.299 & 0.058 & 0.334 & 0.216 & 0.01 & 0.277 & 0.253\\
3 & Chinese & 0.051 & 0.357 & 0.299 & 0.058 & 0.332 & 0.214 & 0.012 & 0.278 & 0.254\\
3 & Czech & 0.051 & 0.357 & 0.299 & 0.058 & 0.333 & 0.215 & 0.01 & 0.279 & 0.254\\
3 & English & 0.051 & 0.357 & 0.299 & 0.067 & 0.357 & 0.24 & 0.009 & 0.275 & 0.254\\
3 & Finnish & 0.051 & 0.357 & 0.299 & 0.06 & 0.337 & 0.219 & 0.013 & 0.287 & 0.26\\
3 & French & 0.051 & 0.357 & 0.299 & 0.066 & 0.354 & 0.237 & 0.012 & 0.276 & 0.251\\
3 & Galician & 0.051 & 0.357 & 0.299 & 0.06 & 0.338 & 0.22 & 0.016 & 0.289 & 0.259\\
3 & German & 0.051 & 0.357 & 0.299 & 0.066 & 0.354 & 0.237 & 0.007 & 0.273 & 0.25\\
3 & Hindi & 0.051 & 0.357 & 0.299 & 0.066 & 0.353 & 0.237 & 0.01 & 0.272 & 0.25\\
3 & Icelandic & 0.051 & 0.357 & 0.299 & 0.066 & 0.354 & 0.237 & 0.01 & 0.274 & 0.25\\
3 & Indonesian & 0.051 & 0.357 & 0.299 & 0.057 & 0.33 & 0.211 & 0.012 & 0.274 & 0.25\\
3 & Italian & 0.051 & 0.357 & 0.299 & 0.058 & 0.333 & 0.215 & 0.009 & 0.279 & 0.255\\
3 & Japanese & 0.051 & 0.357 & 0.299 & 0.061 & 0.34 & 0.223 & 0.009 & 0.257 & 0.238\\
3 & Korean & 0.051 & 0.357 & 0.299 & 0.057 & 0.33 & 0.212 & 0.01 & 0.274 & 0.251\\
3 & Polish & 0.051 & 0.357 & 0.299 & 0.057 & 0.329 & 0.211 & 0.014 & 0.277 & 0.248\\
3 & Portuguese & 0.051 & 0.357 & 0.299 & 0.057 & 0.329 & 0.211 & 0.014 & 0.277 & 0.248\\
3 & Russian & 0.051 & 0.357 & 0.299 & 0.057 & 0.331 & 0.213 & 0.013 & 0.276 & 0.251\\
3 & Spanish & 0.051 & 0.357 & 0.299 & 0.059 & 0.335 & 0.217 & 0.011 & 0.28 & 0.255\\
3 & Swedish & 0.051 & 0.357 & 0.299 & 0.057 & 0.331 & 0.212 & 0.007 & 0.273 & 0.252\\
3 & Thai & 0.051 & 0.357 & 0.299 & 0.056 & 0.328 & 0.21 & 0.005 & 0.27 & 0.248\\
3 & Turkish & 0.051 & 0.357 & 0.299 & 0.059 & 0.335 & 0.217 & 0.015 & 0.283 & 0.255\\

    \end{tabular}
    }
\end{table}

}

\appendixsection{Performance in PUD without sentence length limit}
\label{app:performance_of_random_parser_without_sentence_length_limit}

\begin{table}
    \centering
    \caption{\label{tab:performance_by_score_PUD_UD} The performance scores $\E[P_c^t]$, $\E[Q]$, $\E[P_c^e]$ on human languages in the PUD collection
    \iftoggle{PSUD}{with UD annotation style and}{with} 
    no limit on sentence length. 
    Format is the same as in Table \ref{tab:performance_by_score_PUD10_UD}.
    }    
    {\footnotesize
    \begin{tabular}{rlrrrrrrrrr}
     & & \multicolumn{3}{c}{$\E[P_c^t]$} & \multicolumn{3}{c}{$\E[Q]$} & \multicolumn{3}{c}{$\E[P_c^e]$} \\\cmidrule(lr){3-5} \cmidrule(lr){6-8} \cmidrule(lr){9-11}
    $n_{min}$ & language & u & g & e & u & g & e & u & g & e \\
    \hline
    2 & Arabic & 0.029 & 0.079 & 0.004 & 0.143 & 0.246 & 0.136 & 0.074 & 0.1 & 0.106\\
2 & Chinese & 0.029 & 0.079 & 0.002 & 0.145 & 0.246 & 0.132 & 0.076 & 0.1 & 0.107\\
2 & Czech & 0.033 & 0.092 & 0.004 & 0.157 & 0.272 & 0.155 & 0.084 & 0.115 & 0.123\\
2 & English & 0.026 & 0.082 & 0.001 & 0.131 & 0.254 & 0.129 & 0.067 & 0.104 & 0.106\\
2 & Finnish & 0.037 & 0.109 & 0.005 & 0.171 & 0.304 & 0.181 & 0.094 & 0.136 & 0.146\\
2 & French & 0.027 & 0.068 & 0.001 & 0.135 & 0.225 & 0.111 & 0.069 & 0.088 & 0.089\\
2 & Galician & 0.021 & 0.069 & 0.004 & 0.115 & 0.226 & 0.12 & 0.056 & 0.088 & 0.094\\
2 & German & 0.029 & 0.083 & 0.001 & 0.143 & 0.255 & 0.131 & 0.074 & 0.105 & 0.106\\
2 & Hindi & 0.025 & 0.071 & 0.001 & 0.128 & 0.23 & 0.114 & 0.065 & 0.091 & 0.092\\
2 & Icelandic & 0.028 & 0.092 & 0.002 & 0.139 & 0.273 & 0.145 & 0.072 & 0.116 & 0.117\\
2 & Indonesian & 0.032 & 0.086 & 0.003 & 0.154 & 0.26 & 0.143 & 0.082 & 0.108 & 0.116\\
2 & Italian & 0.024 & 0.068 & 0.002 & 0.125 & 0.223 & 0.117 & 0.063 & 0.087 & 0.092\\
2 & Japanese & 0.02 & 0.058 & 0 & 0.111 & 0.203 & 0.093 & 0.054 & 0.076 & 0.077\\
2 & Korean & 0.033 & 0.098 & 0.004 & 0.16 & 0.283 & 0.162 & 0.086 & 0.123 & 0.131\\
2 & Polish & 0.025 & 0.07 & 0.003 & 0.128 & 0.228 & 0.12 & 0.065 & 0.089 & 0.095\\
2 & Portuguese & 0.025 & 0.07 & 0.003 & 0.128 & 0.228 & 0.12 & 0.065 & 0.089 & 0.095\\
2 & Russian & 0.031 & 0.089 & 0.005 & 0.149 & 0.267 & 0.151 & 0.079 & 0.113 & 0.12\\
2 & Spanish & 0.025 & 0.069 & 0.003 & 0.128 & 0.227 & 0.119 & 0.065 & 0.089 & 0.094\\
2 & Swedish & 0.029 & 0.085 & 0.002 & 0.145 & 0.259 & 0.142 & 0.076 & 0.108 & 0.115\\
2 & Thai & 0.023 & 0.066 & 0.001 & 0.12 & 0.219 & 0.111 & 0.06 & 0.084 & 0.09\\
2 & Turkish & 0.039 & 0.1 & 0.007 & 0.178 & 0.287 & 0.167 & 0.098 & 0.125 & 0.133\\
3 & Arabic & 0.008 & 0.023 & 0.001 & 0.125 & 0.201 & 0.133 & 0.074 & 0.097 & 0.106\\
3 & Chinese & 0.009 & 0.023 & 0.001 & 0.127 & 0.201 & 0.132 & 0.075 & 0.097 & 0.107\\
3 & Czech & 0.01 & 0.027 & 0.002 & 0.137 & 0.22 & 0.153 & 0.083 & 0.111 & 0.123\\
3 & English & 0.007 & 0.024 & 0.001 & 0.115 & 0.206 & 0.129 & 0.066 & 0.101 & 0.106\\
3 & Finnish & 0.011 & 0.032 & 0.004 & 0.149 & 0.243 & 0.18 & 0.093 & 0.13 & 0.146\\
3 & French & 0.008 & 0.02 & 0.001 & 0.118 & 0.185 & 0.111 & 0.068 & 0.086 & 0.089\\
3 & Galician & 0.006 & 0.02 & 0.002 & 0.101 & 0.186 & 0.118 & 0.056 & 0.086 & 0.094\\
3 & German & 0.008 & 0.024 & 0.001 & 0.125 & 0.207 & 0.131 & 0.074 & 0.102 & 0.106\\
3 & Hindi & 0.007 & 0.02 & 0.001 & 0.112 & 0.188 & 0.114 & 0.064 & 0.088 & 0.092\\
3 & Icelandic & 0.008 & 0.027 & 0.002 & 0.122 & 0.22 & 0.145 & 0.071 & 0.112 & 0.117\\
3 & Indonesian & 0.009 & 0.025 & 0.002 & 0.135 & 0.211 & 0.142 & 0.081 & 0.105 & 0.116\\
3 & Italian & 0.007 & 0.02 & 0.001 & 0.11 & 0.184 & 0.116 & 0.062 & 0.085 & 0.092\\
3 & Japanese & 0.006 & 0.017 & 0 & 0.098 & 0.168 & 0.093 & 0.054 & 0.074 & 0.077\\
3 & Korean & 0.01 & 0.028 & 0.003 & 0.139 & 0.228 & 0.161 & 0.085 & 0.118 & 0.131\\
3 & Polish & 0.007 & 0.02 & 0.002 & 0.112 & 0.187 & 0.119 & 0.064 & 0.087 & 0.095\\
3 & Portuguese & 0.007 & 0.02 & 0.002 & 0.112 & 0.187 & 0.119 & 0.064 & 0.087 & 0.095\\
3 & Russian & 0.009 & 0.026 & 0.003 & 0.131 & 0.216 & 0.149 & 0.078 & 0.109 & 0.12\\
3 & Spanish & 0.007 & 0.02 & 0.001 & 0.112 & 0.186 & 0.117 & 0.064 & 0.086 & 0.094\\
3 & Swedish & 0.009 & 0.025 & 0.001 & 0.127 & 0.211 & 0.142 & 0.075 & 0.104 & 0.115\\
3 & Thai & 0.007 & 0.019 & 0 & 0.106 & 0.18 & 0.11 & 0.059 & 0.082 & 0.09\\
3 & Turkish & 0.012 & 0.029 & 0.004 & 0.154 & 0.231 & 0.164 & 0.097 & 0.12 & 0.133\\

    \end{tabular}
    }
    
\end{table}

\begin{table}
    \centering
\caption{\label{tab:performance_by_distribution_PUD_UD}The performance scores $\E[P_c^t]$, $\E[Q]$, $\E[P_c^e]$ on human languages in the PUD collection 
\iftoggle{PSUD}{with UD annotation style and}{with} no limit on sentence length. Format is the same as in Table \ref{tab:performance_by_distribution_PUD10_UD}.
    }    
    {\footnotesize
    \begin{tabular}{rlrrrrrrrrr}
     & & \multicolumn{3}{c}{uniform} & \multicolumn{3}{c}{geometric} & \multicolumn{3}{c}{empirical} \\\cmidrule(lr){3-5} \cmidrule(lr){6-8} \cmidrule(lr){9-11}
    $n_{min}$ & language &  
    $\E[P_c^t]$ & $\E[Q]$ & $\E[P_c^e]$ &   
    $\E[P_c^t]$ & $\E[Q]$ & $\E[P_c^e]$ &
    $\E[P_c^t]$ & $\E[Q]$ & $\E[P_c^e]$ \\
    \hline
    2 & Arabic & 0.029 & 0.143 & 0.074 & 0.079 & 0.246 & 0.1 & 0.004 & 0.136 & 0.106\\
2 & Chinese & 0.029 & 0.145 & 0.076 & 0.079 & 0.246 & 0.1 & 0.002 & 0.132 & 0.107\\
2 & Czech & 0.033 & 0.157 & 0.084 & 0.092 & 0.272 & 0.115 & 0.004 & 0.155 & 0.123\\
2 & English & 0.026 & 0.131 & 0.067 & 0.082 & 0.254 & 0.104 & 0.001 & 0.129 & 0.106\\
2 & Finnish & 0.037 & 0.171 & 0.094 & 0.109 & 0.304 & 0.136 & 0.005 & 0.181 & 0.146\\
2 & French & 0.027 & 0.135 & 0.069 & 0.068 & 0.225 & 0.088 & 0.001 & 0.111 & 0.089\\
2 & Galician & 0.021 & 0.115 & 0.056 & 0.069 & 0.226 & 0.088 & 0.004 & 0.12 & 0.094\\
2 & German & 0.029 & 0.143 & 0.074 & 0.083 & 0.255 & 0.105 & 0.001 & 0.131 & 0.106\\
2 & Hindi & 0.025 & 0.128 & 0.065 & 0.071 & 0.23 & 0.091 & 0.001 & 0.114 & 0.092\\
2 & Icelandic & 0.028 & 0.139 & 0.072 & 0.092 & 0.273 & 0.116 & 0.002 & 0.145 & 0.117\\
2 & Indonesian & 0.032 & 0.154 & 0.082 & 0.086 & 0.26 & 0.108 & 0.003 & 0.143 & 0.116\\
2 & Italian & 0.024 & 0.125 & 0.063 & 0.068 & 0.223 & 0.087 & 0.002 & 0.117 & 0.092\\
2 & Japanese & 0.02 & 0.111 & 0.054 & 0.058 & 0.203 & 0.076 & 0 & 0.093 & 0.077\\
2 & Korean & 0.033 & 0.16 & 0.086 & 0.098 & 0.283 & 0.123 & 0.004 & 0.162 & 0.131\\
2 & Polish & 0.025 & 0.128 & 0.065 & 0.07 & 0.228 & 0.089 & 0.003 & 0.12 & 0.095\\
2 & Portuguese & 0.025 & 0.128 & 0.065 & 0.07 & 0.228 & 0.089 & 0.003 & 0.12 & 0.095\\
2 & Russian & 0.031 & 0.149 & 0.079 & 0.089 & 0.267 & 0.113 & 0.005 & 0.151 & 0.12\\
2 & Spanish & 0.025 & 0.128 & 0.065 & 0.069 & 0.227 & 0.089 & 0.003 & 0.119 & 0.094\\
2 & Swedish & 0.029 & 0.145 & 0.076 & 0.085 & 0.259 & 0.108 & 0.002 & 0.142 & 0.115\\
2 & Thai & 0.023 & 0.12 & 0.06 & 0.066 & 0.219 & 0.084 & 0.001 & 0.111 & 0.09\\
2 & Turkish & 0.039 & 0.178 & 0.098 & 0.1 & 0.287 & 0.125 & 0.007 & 0.167 & 0.133\\
3 & Arabic & 0.008 & 0.125 & 0.074 & 0.023 & 0.201 & 0.097 & 0.001 & 0.133 & 0.106\\
3 & Chinese & 0.009 & 0.127 & 0.075 & 0.023 & 0.201 & 0.097 & 0.001 & 0.132 & 0.107\\
3 & Czech & 0.01 & 0.137 & 0.083 & 0.027 & 0.22 & 0.111 & 0.002 & 0.153 & 0.123\\
3 & English & 0.007 & 0.115 & 0.066 & 0.024 & 0.206 & 0.101 & 0.001 & 0.129 & 0.106\\
3 & Finnish & 0.011 & 0.149 & 0.093 & 0.032 & 0.243 & 0.13 & 0.004 & 0.18 & 0.146\\
3 & French & 0.008 & 0.118 & 0.068 & 0.02 & 0.185 & 0.086 & 0.001 & 0.111 & 0.089\\
3 & Galician & 0.006 & 0.101 & 0.056 & 0.02 & 0.186 & 0.086 & 0.002 & 0.118 & 0.094\\
3 & German & 0.008 & 0.125 & 0.074 & 0.024 & 0.207 & 0.102 & 0.001 & 0.131 & 0.106\\
3 & Hindi & 0.007 & 0.112 & 0.064 & 0.02 & 0.188 & 0.088 & 0.001 & 0.114 & 0.092\\
3 & Icelandic & 0.008 & 0.122 & 0.071 & 0.027 & 0.22 & 0.112 & 0.002 & 0.145 & 0.117\\
3 & Indonesian & 0.009 & 0.135 & 0.081 & 0.025 & 0.211 & 0.105 & 0.002 & 0.142 & 0.116\\
3 & Italian & 0.007 & 0.11 & 0.062 & 0.02 & 0.184 & 0.085 & 0.001 & 0.116 & 0.092\\
3 & Japanese & 0.006 & 0.098 & 0.054 & 0.017 & 0.168 & 0.074 & 0 & 0.093 & 0.077\\
3 & Korean & 0.01 & 0.139 & 0.085 & 0.028 & 0.228 & 0.118 & 0.003 & 0.161 & 0.131\\
3 & Polish & 0.007 & 0.112 & 0.064 & 0.02 & 0.187 & 0.087 & 0.002 & 0.119 & 0.095\\
3 & Portuguese & 0.007 & 0.112 & 0.064 & 0.02 & 0.187 & 0.087 & 0.002 & 0.119 & 0.095\\
3 & Russian & 0.009 & 0.131 & 0.078 & 0.026 & 0.216 & 0.109 & 0.003 & 0.149 & 0.12\\
3 & Spanish & 0.007 & 0.112 & 0.064 & 0.02 & 0.186 & 0.086 & 0.001 & 0.117 & 0.094\\
3 & Swedish & 0.009 & 0.127 & 0.075 & 0.025 & 0.211 & 0.104 & 0.001 & 0.142 & 0.115\\
3 & Thai & 0.007 & 0.106 & 0.059 & 0.019 & 0.18 & 0.082 & 0 & 0.11 & 0.09\\
3 & Turkish & 0.012 & 0.154 & 0.097 & 0.029 & 0.231 & 0.12 & 0.004 & 0.164 & 0.133\\

    \end{tabular}
    }    
\end{table}

\iftoggle{PSUD}{
\begin{table}
    \centering
    \caption{\label{tab:performance_by_score_PUD_SUD}The performance scores $\E[P_c^t]$, $\E[Q]$, $\E[P_c^e]$ on human languages in the PUD collection with SUD annotation style and no limit on sentence length. 
    Format is the same as in Table \ref{tab:performance_by_score_PUD10_UD}.}    
    {\footnotesize
    \begin{tabular}{rlrrrrrrrrr}
     & & \multicolumn{3}{c}{$\E[P_c^t]$} & \multicolumn{3}{c}{$\E[Q]$} & \multicolumn{3}{c}{$\E[P_c^e]$} \\\cmidrule(lr){3-5} \cmidrule(lr){6-8} \cmidrule(lr){9-11}
    $n_{min}$ & language & u & g & e & u & g & e & u & g & e \\
    \hline
    2 & Arabic & 0.029 & 0.079 & 0.004 & 0.143 & 0.246 & 0.136 & 0.074 & 0.1 & 0.106\\
2 & Chinese & 0.029 & 0.079 & 0.002 & 0.145 & 0.246 & 0.132 & 0.076 & 0.1 & 0.107\\
2 & Czech & 0.033 & 0.092 & 0.004 & 0.157 & 0.272 & 0.155 & 0.084 & 0.115 & 0.123\\
2 & English & 0.026 & 0.082 & 0.001 & 0.131 & 0.254 & 0.129 & 0.067 & 0.104 & 0.106\\
2 & Finnish & 0.037 & 0.109 & 0.005 & 0.171 & 0.304 & 0.181 & 0.094 & 0.136 & 0.146\\
2 & French & 0.027 & 0.068 & 0.001 & 0.135 & 0.225 & 0.111 & 0.069 & 0.088 & 0.089\\
2 & Galician & 0.021 & 0.069 & 0.004 & 0.115 & 0.226 & 0.12 & 0.056 & 0.088 & 0.094\\
2 & German & 0.029 & 0.083 & 0.001 & 0.143 & 0.255 & 0.131 & 0.074 & 0.105 & 0.106\\
2 & Hindi & 0.025 & 0.071 & 0.001 & 0.128 & 0.23 & 0.114 & 0.065 & 0.091 & 0.092\\
2 & Icelandic & 0.028 & 0.092 & 0.002 & 0.139 & 0.273 & 0.145 & 0.072 & 0.116 & 0.117\\
2 & Indonesian & 0.032 & 0.086 & 0.003 & 0.154 & 0.26 & 0.143 & 0.082 & 0.108 & 0.116\\
2 & Italian & 0.024 & 0.068 & 0.002 & 0.125 & 0.223 & 0.117 & 0.063 & 0.087 & 0.092\\
2 & Japanese & 0.02 & 0.058 & 0 & 0.111 & 0.203 & 0.093 & 0.054 & 0.076 & 0.077\\
2 & Korean & 0.033 & 0.098 & 0.004 & 0.16 & 0.283 & 0.162 & 0.086 & 0.123 & 0.131\\
2 & Polish & 0.025 & 0.07 & 0.003 & 0.128 & 0.228 & 0.12 & 0.065 & 0.089 & 0.095\\
2 & Portuguese & 0.025 & 0.07 & 0.003 & 0.128 & 0.228 & 0.12 & 0.065 & 0.089 & 0.095\\
2 & Russian & 0.031 & 0.089 & 0.005 & 0.149 & 0.267 & 0.151 & 0.079 & 0.113 & 0.12\\
2 & Spanish & 0.025 & 0.069 & 0.003 & 0.128 & 0.227 & 0.119 & 0.065 & 0.089 & 0.094\\
2 & Swedish & 0.029 & 0.085 & 0.002 & 0.145 & 0.259 & 0.142 & 0.076 & 0.108 & 0.115\\
2 & Thai & 0.023 & 0.066 & 0.001 & 0.12 & 0.219 & 0.111 & 0.06 & 0.084 & 0.09\\
2 & Turkish & 0.039 & 0.1 & 0.007 & 0.178 & 0.287 & 0.167 & 0.098 & 0.125 & 0.133\\
3 & Arabic & 0.008 & 0.023 & 0.001 & 0.125 & 0.201 & 0.133 & 0.074 & 0.097 & 0.106\\
3 & Chinese & 0.009 & 0.023 & 0.001 & 0.127 & 0.201 & 0.132 & 0.075 & 0.097 & 0.107\\
3 & Czech & 0.01 & 0.027 & 0.002 & 0.137 & 0.22 & 0.153 & 0.083 & 0.111 & 0.123\\
3 & English & 0.007 & 0.024 & 0.001 & 0.115 & 0.206 & 0.129 & 0.066 & 0.101 & 0.106\\
3 & Finnish & 0.011 & 0.032 & 0.004 & 0.149 & 0.243 & 0.18 & 0.093 & 0.13 & 0.146\\
3 & French & 0.008 & 0.02 & 0.001 & 0.118 & 0.185 & 0.111 & 0.068 & 0.086 & 0.089\\
3 & Galician & 0.006 & 0.02 & 0.002 & 0.101 & 0.186 & 0.118 & 0.056 & 0.086 & 0.094\\
3 & German & 0.008 & 0.024 & 0.001 & 0.125 & 0.207 & 0.131 & 0.074 & 0.102 & 0.106\\
3 & Hindi & 0.007 & 0.02 & 0.001 & 0.112 & 0.188 & 0.114 & 0.064 & 0.088 & 0.092\\
3 & Icelandic & 0.008 & 0.027 & 0.002 & 0.122 & 0.22 & 0.145 & 0.071 & 0.112 & 0.117\\
3 & Indonesian & 0.009 & 0.025 & 0.002 & 0.135 & 0.211 & 0.142 & 0.081 & 0.105 & 0.116\\
3 & Italian & 0.007 & 0.02 & 0.001 & 0.11 & 0.184 & 0.116 & 0.062 & 0.085 & 0.092\\
3 & Japanese & 0.006 & 0.017 & 0 & 0.098 & 0.168 & 0.093 & 0.054 & 0.074 & 0.077\\
3 & Korean & 0.01 & 0.028 & 0.003 & 0.139 & 0.228 & 0.161 & 0.085 & 0.118 & 0.131\\
3 & Polish & 0.007 & 0.02 & 0.002 & 0.112 & 0.187 & 0.119 & 0.064 & 0.087 & 0.095\\
3 & Portuguese & 0.007 & 0.02 & 0.002 & 0.112 & 0.187 & 0.119 & 0.064 & 0.087 & 0.095\\
3 & Russian & 0.009 & 0.026 & 0.003 & 0.131 & 0.216 & 0.149 & 0.078 & 0.109 & 0.12\\
3 & Spanish & 0.007 & 0.02 & 0.001 & 0.112 & 0.186 & 0.117 & 0.064 & 0.086 & 0.094\\
3 & Swedish & 0.009 & 0.025 & 0.001 & 0.127 & 0.211 & 0.142 & 0.075 & 0.104 & 0.115\\
3 & Thai & 0.007 & 0.019 & 0 & 0.106 & 0.18 & 0.11 & 0.059 & 0.082 & 0.09\\
3 & Turkish & 0.012 & 0.029 & 0.004 & 0.154 & 0.231 & 0.164 & 0.097 & 0.12 & 0.133\\
    
    \end{tabular}
    }
\end{table}

\begin{table}
    \centering
    \caption{\label{tab:performance_by_distribution_PUD_SUD}The performance scores $\E[P_c^t]$, $\E[Q]$, $\E[P_c^e]$ on human languages in the PUD collection with SUD annotation style and no limit on sentence length. Format is the same as in Table \ref{tab:performance_by_distribution_PUD10_UD}.}    
    {\footnotesize
    \begin{tabular}{rlrrrrrrrrr}
     & & \multicolumn{3}{c}{uniform} & \multicolumn{3}{c}{geometric} & \multicolumn{3}{c}{empirical} \\\cmidrule(lr){3-5} \cmidrule(lr){6-8} \cmidrule(lr){9-11}
     $n_{min}$ & language & $\E[P_c^t]$ & $\E[Q]$ & $\E[P_c^e]$ & $\E[P_c^t]$ & $\E[Q]$ & $\E[P_c^e]$ & $\E[P_c^t]$ & $\E[Q]$ & $\E[P_c^e]$ \\
    \hline
    2 & Arabic & 0.029 & 0.143 & 0.074 & 0.079 & 0.246 & 0.1 & 0.004 & 0.136 & 0.106\\
2 & Chinese & 0.029 & 0.145 & 0.076 & 0.079 & 0.246 & 0.1 & 0.002 & 0.132 & 0.107\\
2 & Czech & 0.033 & 0.157 & 0.084 & 0.092 & 0.272 & 0.115 & 0.004 & 0.155 & 0.123\\
2 & English & 0.026 & 0.131 & 0.067 & 0.082 & 0.254 & 0.104 & 0.001 & 0.129 & 0.106\\
2 & Finnish & 0.037 & 0.171 & 0.094 & 0.109 & 0.304 & 0.136 & 0.005 & 0.181 & 0.146\\
2 & French & 0.027 & 0.135 & 0.069 & 0.068 & 0.225 & 0.088 & 0.001 & 0.111 & 0.089\\
2 & Galician & 0.021 & 0.115 & 0.056 & 0.069 & 0.226 & 0.088 & 0.004 & 0.12 & 0.094\\
2 & German & 0.029 & 0.143 & 0.074 & 0.083 & 0.255 & 0.105 & 0.001 & 0.131 & 0.106\\
2 & Hindi & 0.025 & 0.128 & 0.065 & 0.071 & 0.23 & 0.091 & 0.001 & 0.114 & 0.092\\
2 & Icelandic & 0.028 & 0.139 & 0.072 & 0.092 & 0.273 & 0.116 & 0.002 & 0.145 & 0.117\\
2 & Indonesian & 0.032 & 0.154 & 0.082 & 0.086 & 0.26 & 0.108 & 0.003 & 0.143 & 0.116\\
2 & Italian & 0.024 & 0.125 & 0.063 & 0.068 & 0.223 & 0.087 & 0.002 & 0.117 & 0.092\\
2 & Japanese & 0.02 & 0.111 & 0.054 & 0.058 & 0.203 & 0.076 & 0 & 0.093 & 0.077\\
2 & Korean & 0.033 & 0.16 & 0.086 & 0.098 & 0.283 & 0.123 & 0.004 & 0.162 & 0.131\\
2 & Polish & 0.025 & 0.128 & 0.065 & 0.07 & 0.228 & 0.089 & 0.003 & 0.12 & 0.095\\
2 & Portuguese & 0.025 & 0.128 & 0.065 & 0.07 & 0.228 & 0.089 & 0.003 & 0.12 & 0.095\\
2 & Russian & 0.031 & 0.149 & 0.079 & 0.089 & 0.267 & 0.113 & 0.005 & 0.151 & 0.12\\
2 & Spanish & 0.025 & 0.128 & 0.065 & 0.069 & 0.227 & 0.089 & 0.003 & 0.119 & 0.094\\
2 & Swedish & 0.029 & 0.145 & 0.076 & 0.085 & 0.259 & 0.108 & 0.002 & 0.142 & 0.115\\
2 & Thai & 0.023 & 0.12 & 0.06 & 0.066 & 0.219 & 0.084 & 0.001 & 0.111 & 0.09\\
2 & Turkish & 0.039 & 0.178 & 0.098 & 0.1 & 0.287 & 0.125 & 0.007 & 0.167 & 0.133\\
3 & Arabic & 0.008 & 0.125 & 0.074 & 0.023 & 0.201 & 0.097 & 0.001 & 0.133 & 0.106\\
3 & Chinese & 0.009 & 0.127 & 0.075 & 0.023 & 0.201 & 0.097 & 0.001 & 0.132 & 0.107\\
3 & Czech & 0.01 & 0.137 & 0.083 & 0.027 & 0.22 & 0.111 & 0.002 & 0.153 & 0.123\\
3 & English & 0.007 & 0.115 & 0.066 & 0.024 & 0.206 & 0.101 & 0.001 & 0.129 & 0.106\\
3 & Finnish & 0.011 & 0.149 & 0.093 & 0.032 & 0.243 & 0.13 & 0.004 & 0.18 & 0.146\\
3 & French & 0.008 & 0.118 & 0.068 & 0.02 & 0.185 & 0.086 & 0.001 & 0.111 & 0.089\\
3 & Galician & 0.006 & 0.101 & 0.056 & 0.02 & 0.186 & 0.086 & 0.002 & 0.118 & 0.094\\
3 & German & 0.008 & 0.125 & 0.074 & 0.024 & 0.207 & 0.102 & 0.001 & 0.131 & 0.106\\
3 & Hindi & 0.007 & 0.112 & 0.064 & 0.02 & 0.188 & 0.088 & 0.001 & 0.114 & 0.092\\
3 & Icelandic & 0.008 & 0.122 & 0.071 & 0.027 & 0.22 & 0.112 & 0.002 & 0.145 & 0.117\\
3 & Indonesian & 0.009 & 0.135 & 0.081 & 0.025 & 0.211 & 0.105 & 0.002 & 0.142 & 0.116\\
3 & Italian & 0.007 & 0.11 & 0.062 & 0.02 & 0.184 & 0.085 & 0.001 & 0.116 & 0.092\\
3 & Japanese & 0.006 & 0.098 & 0.054 & 0.017 & 0.168 & 0.074 & 0 & 0.093 & 0.077\\
3 & Korean & 0.01 & 0.139 & 0.085 & 0.028 & 0.228 & 0.118 & 0.003 & 0.161 & 0.131\\
3 & Polish & 0.007 & 0.112 & 0.064 & 0.02 & 0.187 & 0.087 & 0.002 & 0.119 & 0.095\\
3 & Portuguese & 0.007 & 0.112 & 0.064 & 0.02 & 0.187 & 0.087 & 0.002 & 0.119 & 0.095\\
3 & Russian & 0.009 & 0.131 & 0.078 & 0.026 & 0.216 & 0.109 & 0.003 & 0.149 & 0.12\\
3 & Spanish & 0.007 & 0.112 & 0.064 & 0.02 & 0.186 & 0.086 & 0.001 & 0.117 & 0.094\\
3 & Swedish & 0.009 & 0.127 & 0.075 & 0.025 & 0.211 & 0.104 & 0.001 & 0.142 & 0.115\\
3 & Thai & 0.007 & 0.106 & 0.059 & 0.019 & 0.18 & 0.082 & 0 & 0.11 & 0.09\\
3 & Turkish & 0.012 & 0.154 & 0.097 & 0.029 & 0.231 & 0.12 & 0.004 & 0.164 & 0.133\\

    \end{tabular}
    }    
\end{table}
}


\begin{acknowledgments}
We are grateful to L. Màrquez and L. Padró for helpful discussions. We are also grateful to M. Mora for advice on mathematical calculations and to L. Alemany-Puig for assistance in the generation of the preprocessed PUD treebanks. RFC is supported by the grant PID2024-155946NB-I00 funded by Ministerio de Ciencia, Innovación y Universidades (MICIU), Agencia Estatal de Investigación (AEI/10.13039/501100011033) and the European Social Fund Plus (ESF+).
CH is supported by the grant 802719 funded by the European Union’s 8th Framework Programme (Horizon 2020) and by the gran TWCF-2024-33965 from the Templeton World Charity Foundation.
\end{acknowledgments}


\clearpage

\bibliographystyle{compling}
\bibliography{biblio.bib,new_references.bib}

\end{document}